\documentclass[pdflatex, sn-apa, iicol]{sn-jnl}

\usepackage{graphicx}%
\usepackage{multirow}%
\usepackage{amsmath,amssymb,amsfonts, dsfont}%
\usepackage{amsthm}%
\usepackage{mathrsfs}%
\usepackage{mathtools}
\usepackage[title]{appendix}%
\usepackage[table,usenames,dvipsnames]{xcolor}%
\usepackage{textcomp}%
\usepackage{manyfoot}%
\usepackage{booktabs}%
\usepackage{algorithm}%
\usepackage{algorithmicx}%
\usepackage{algpseudocode}%
\usepackage{listings}%
\usepackage[export]{adjustbox}
\usepackage{graphicx,tabularx,subcaption}

\theoremstyle{thmstyleone}%
\newtheorem{theorem}{Theorem}
\newtheorem{proposition}[theorem]{Proposition}%

\theoremstyle{thmstyletwo}%
\newtheorem{remark}{Remark}%

\theoremstyle{thmstylethree}%
\newtheorem{definition}{Definition}%
\newtheorem{assumption}{Assumption}%

\newcommand{\calA}{\mathcal{A}}
\newcommand{\calB}{\mathcal{B}}

\newcommand{\calI}{\mathcal{I}}

\newcommand{\calK}{\mathcal{K}}

\newcommand{\calM}{\mathcal{M}}

\newcommand{\calO}{\mathcal{O}}
\newcommand{\calP}{\mathcal{P}}

\newcommand{\calR}{\mathcal{R}}

\newcommand{\calW}{\mathcal{W}}

\newcommand{\bfa}{\mathbf{a}}
\newcommand{\bfb}{\mathbf{b}}

\newcommand{\bfe}{\mathbf{e}}
\newcommand{\bff}{\mathbf{f}}
\newcommand{\bfg}{\mathbf{g}}

\newcommand{\bfk}{\mathbf{k}}

\newcommand{\bfp}{\mathbf{p}}
\newcommand{\bfq}{\mathbf{q}}

\newcommand{\bfs}{\mathbf{s}}

\newcommand{\bfu}{\mathbf{u}}
\newcommand{\bfv}{\mathbf{v}}

\newcommand{\bfx}{\mathbf{x}}
\newcommand{\bfy}{\mathbf{y}}
\newcommand{\bfz}{\mathbf{z}}

\newcommand{\bfI}{\mathbf{I}}

\newcommand{\bfQ}{\mathbf{Q}}

\newcommand{\bbR}{\mathbb{R}}

\newcommand{\prl}[1]{\left(#1\right)}
\newcommand{\brl}[1]{\left[#1\right]}
\newcommand{\crl}[1]{\left\{#1\right\}}

\DeclarePairedDelimiterX{\norm}[1]{\lVert}{\rVert}{#1}
\DeclarePairedDelimiterX{\abs}[1]{\mid}{\mid}{#1}

\newcommand*{\Int}{\text{int}}
\newcommand*{\Cl}{\text{cl}}

\newcommand*{\LS}{\mathcal{LS}}

\newcommand*{\carspeed}{v}

\newcommand*{\carorientation}{\theta}

\begin{document}

\title{EAST: Environment-Aware Safe Tracking for Robot Navigation in Dynamic Environments}


\author[1]{\fnm{Zhichao} \sur{Li}}\email{zhichaoli@ucsd.edu} 
\equalcont{These authors contributed equally to this work.}

\author*[1]{\fnm{Yinzhuang} \sur{Yi}}\email{yiyi@ucsd.edu}
\equalcont{These authors contributed equally to this work.}

\author[1]{\fnm{Zhuolin} \sur{Niu}}\email{zhniu@ucsd.edu}

\author[1]{\fnm{Nikolay} \sur{Atanasov}}\email{natanasov@ucsd.edu}

\affil[1]{\orgdiv{Department of Electrical and Computer Engineering}, \orgname{University of California San Diego}, \orgaddress{\city{La Jolla}, \postcode{92093}, \state{CA}, \country{USA}}}


\abstract{This paper considers the problem of autonomous mobile robot navigation in unknown environments with moving obstacles.  We propose a new method to achieve environment-aware safe tracking (EAST) of robot motion plans that integrates an obstacle clearance cost for path planning, a convex reachable set for robot motion prediction, and safety constraints for dynamic obstacle avoidance. EAST adapts the motion of the robot according to the locally sensed environment geometry and dynamics, leading to fast motion in wide open areas and cautious behavior in narrow passages or near moving obstacles. Our control design uses a reference governor, a virtual dynamical system that guides the robot's motion and decouples the path tracking and safety objectives. While reference governor methods have been used for safe tracking control in static environments, our key contribution is an extension to dynamic environments using convex optimization with control barrier function (CBF) constraints. Thus, our work establishes a connection between reference governor techniques and CBF techniques for safe control in dynamic environments. We validate our approach in simulated and real-world environments, featuring complex obstacle configurations and natural dynamic obstacle motion.}




\maketitle

\section{Introduction}
\label{sec:introduction}

Autonomous mobile robots are being increasingly integrated into human environments to support services including transportation, infrastructure inspection, cleaning, and medical assistance. Reliable robot navigation is a crucial aspect for enabling these services, leading to a growing focus on developing certifiably safe yet efficient robot navigation techniques. 

In this paper, we consider the problem of safe robot navigation in unknown environments with moving obstacles. The objective is to enable a mobile robot to navigate to a desired goal safely, potentially without prior knowledge of the environment, relying solely on onboard sensing.

The safe control problem can be approached using model predictive control (MPC) \citep{morari2017predictive, MPC_tube_gao2014tube, MPC_CBF_MAS_Jankovic2020, MPC_Automatica_Camacho2006_nlin, saccani2023mpc, wen2024mpc, dircks2025mpc}, quadratic programming with control barrier function (CBF) constraints \citep{CBF_ames2019ECC, CBF_Ames2015_swarm, CBF_ames2017TAC, CBF_magnus2018CA_CBF, dai2023cbf, allibhoy2024cbf, mestres2024dcbf, chandra2025cbf}, or Hamilton–Jacobi (HJ) reachability analysis \citep{ding2011hj, HJ_time_varying, margellos2011tv_hj, sharpless2023koopmanhi, bajcsy2019unknown_hj, ganai2024learning_hj}. However, many of these methods require prior knowledge of the obstacle configuration to form safety constraints, and their extension to time-varying safety constraints is non-trivial.

In this paper, we consider the reference governor method \citep{RG_bemporad1998reference, RG_garone2016_ERG, RG_garone2017survey, nicotra2018erg, Li_SafeControl_ICRA20}, which does not impose constraints on the system dynamics directly. Instead, a virtual dynamical system, called \emph{reference governor}, is introduced to decouple the stabilization and safety objectives. The governor state acts as a desired equilibrium point for the robot, while safety constraints are enforced by controlling the evolution of the governor state by comparing the obstacle distance to the robot's Lyapunov function. With few exceptions \citep{hossein2020tv_rg, miguel2024tvrg}, reference governor techniques have been used to achieve safe tracking only in static environments. 

Our \textbf{contribution} is a new reference governor tracking control design based on optimization with Lyapunov function constraints to account for static obtacles and time-varying CBF constraints to account for dynamic obstacles. The resulting optimization is a convex quadratically constrained quadratic program, which can be solved efficiently. Our formulation establishes a connection between reference governor techniques and CBF techniques for safe control synthesis. Based on this formulation, we develop an environment-aware safe tracking (EAST) method that dynamically adjusts the governor state to avoid both static and dynamic obstacles, while guiding the robot along a reference path. EAST also integrates a motion planner with an obstacle clearance cost to periodically replan the reference path for the governor as new observations from the environment are received by the robot. Thus, EAST can guide a mobile robot through an unknown environment, sensing obstacles online and relying on low-frequency planning and high-frequency control to make necessary maneuvers, while providing rigorous stability and safety guarantees. 

We evaluate the performance of EAST extensively in both simulated and real-world environments, featuring large areas with complex obstacle configurations and natural obstacle motion. An open-source implementation of EAST is available at \url{https://github.com/ExistentialRobotics/EAST}.

\section{Related Work}
\label{sec:related_works}

In safety-critical robotics applications, safety requirements are typically formulated as constraints in an optimization problem, aiming to synthesize control inputs or control policies for the robot. These safety requirements may arise from diverse considerations, including geometric constraints, actuator limits \citep{fan2019mid}, or more abstract task-specific and semantic-rich constraints \citep{nakamura2025generalizing}. MPC \citep{morari2017predictive} is a widely used approach for safe control synthesis, which approximates an infinite-horizon optimal control problem with a sequence of finite-horizon problems. To ensure recursive feasibility, MPC formulations typically require the constraint sets to be polytopic \citep{morari2017predictive, MPC_tube_gao2014tube} or the feasible space to be approximated with convex regions \citep{SFC_FM}. The EVA-Planner \citep{eva_planner2021} is an MPC-based method for safe tracking that is closely related to our work. To deal with general constraints, the EVA-Planner incorporates the constraints as (soft) cost terms and uses a two-level MPC formulation proposed in \citep{gao2010predictive}. The high-level MPC uses a reduced-order model to generate a feasible reference trajectory efficiently, while the low-level MPC uses the full-order model to generate control inputs for path tracking. In contrast, our method imposes hard constraints using a reduced-order reference governor system and directly handles general (non-convex) constraints. 

In the presence of dynamic obstacles, the constraints become time-varying. Applying MPC techniques in such settings requires motion prediction of the dynamic obstacles \citep{brito2019mov_obs, lin2020mov_obs}. Our method also requires obstacle motion prediction but can directly handle non-convex constraints encoded as time-varying CBF on the governor system input. 

CBF methods \citep{CBF_ames2014control, CBF_ames2017TAC, CBF_ames2019ECC} are increasingly used in safety-critical control applications, owing to their simplicity for ensuring safety with non-convex and dynamics-aware constraints. As discussed in \citet{CBF_ames2014control}, for control-affine systems, the CBF conditions can be expressed as linear constraints in the control input. This structure enables the synthesis of safe controllers via quadratic programming. However, constructing valid CBFs remains a challenging task \citep{CBF_ames2019ECC}, often requiring prior knowledge of obstacles, which can limit applicability in dynamic or partially known environments. \citet{learn2021long} construct CBF constraints for safe robot navigation online from onboard range sensing. However, this approach is limited to static obstacles. Furthermore, CBF constraints derived from onboard measurements \citep{learn2021long, keyumarsi2024lidar_cbf} typically require the system to have relative degree one. Extending such methods to higher-relative-degree systems requires non-trivial generalizations \citep{CBF_HOCBF_xiao2019control, CBF_ECBF_nguyen2016ACC}. In contrast, our formulation imposes the CBF constraints on the governor system, which is of relative degree one by design, thus circumventing the complexity associated with higher-order CBF constructions.

Reachability-based approaches for safe robot navigation depend on accurate reachable set approximations. Various techniques have been developed to approximate reachable sets, including sum-of-squares optimization \citep{kousik2020bridging_ijrr}, funnels \citep{funnel_sequential_composition_Burridge1999}, and Hamilton-Jacobi (HJ) reachability analysis \citep{ding2011hj, hjr}. Notably, HJ reachability analysis has demonstrated its effectiveness for safe robot navigation \citep{borquez2024safety, herbert2017fastrack, seo2019robust_hj}. Although time-varying constraints can be handled by the HJ reachability approach \citep{HJ_time_varying, margellos2011tv_hj}, the reachability computations require solving HJ partial differential equations, which is challenging for high-dimensional systems. In contrast, our approach enforces safety in dynamic environments by formulating a convex quadratically constrained quadratic program for the reduced-order reference governor system, providing a computationally efficient alternative to HJ reachability methods.

To address the challenge of satisfying safety and stability constraints simultaneously, reference governor techniques \citep{RG_bemporad1998reference, RG_garone2016_ERG, RG_garone2017survey} assume the availability of a pre-defined control law that stabilizes the system to an arbitrary equilibrium. The method introduces a virtual reference governor system to act as an equilibrium for the original system, and safety constraints are imposed on the reference governor, thus decoupling the objectives of stabilization and safety. Reference governor methods have been successfully employed in a variety of applications, notable for ensuring safe aerial robot navigation \citep{constrained2024gaetano, covens2022aerial_erg} and safe ground robot navigation in unknown environments \citep{RG_Omur_ICRA17, Li_GCBF_Automatica23}. \citet{gautam2025compliant} also extended the reference governor method to enforce kinematic and force constraints for robot manipulators, while transitioning between free motion and contact modes. However, existing applications of the reference governor are largely restricted to static environments. \citet{hossein2020tv_rg} extended the explicit reference governor approach of \citep{RG_nicotra2018_ERG} to accommodate time-varying constraints, under the assumption that the evolution of the constraints is known a priori. To address this limitation, \citet{miguel2024tvrg} recast the control problem with time-varying constraints as an equivalent problem with time-invariant constraints and an unmeasured disturbance, thereby enabling the application of robust reference governor techniques \citep{miguel2024robust_rg}. In contrast, our method offers a simple yet effective extension to the reference governor method by incorporating time-varying CBF constraints to account for dynamic obstacles. Our approach formulates a convex quadratically constrained quadratic program to adapt the reference governor input, enabling its application in environments with moving obstacles.

\section{Problem Formulation}
\label{sec:problem_formulation}

Consider a ground wheeled robot with differential-drive kinematics:
\begin{equation} \label{eq:kinematic_unicycle}
\dot{\bfx} = 
\begin{bmatrix}
\dot{x} \\
\dot{y} \\
\dot{\theta} \\
\end{bmatrix} = \begin{bmatrix}
\cos \carorientation & 0 \\
\sin \carorientation & 0\\
0 & 1\\ 
\end{bmatrix} \begin{bmatrix}
\carspeed \\
\omega
\end{bmatrix} = G(\bfx) \bfu,
\end{equation}
where the state $\bfx = (x,y,\theta)$ consists of position $\bfp = (x,y)$ and orientation $\theta$, while the input $\bfu = (v,\omega)$ consists of linear velocity $v$ and angular velocity $\omega$. The robot's body is contained in a Euclidean ball $\calB_{r}(\bfp) \coloneqq \crl{\bfz \in \bbR^2 \mid \norm{\bfz - \bfp} \leq r}$ centered at $\bfp$ with radius $r$.

The robot is operating in an unknown environment $\calW \subset \bbR^2$ containing both static and moving obstacles. The static obstacles are modeled as a closed set $\Omega \subset \calW$. Each moving obstacle $i \in \calI = \{1,2,...,k\}$ is modeled as a ball $\calB_{r_i}(\bfp_i)$ with radius $r_i$, position $\bfp_i$, and velocity $\bfv_i$ such that:
\begin{equation}
    \dot{\bfp}_i = \bfv_i,
\end{equation}
where the velocity can be time varying.

The static obstacle space can be inflated to capture all positions such that the robot body is not contained in the free space of the static environment:
\begin{equation}
    \Omega^+ = \bigcup_{\bfq \in \Omega} \calB_r(\bfq).
\end{equation}
Similarly, define an inflated obstacle space capturing both the static and dynamic obstacles:
\begin{equation}
    \calO^+(t) = \left( \bigcup_{i \in \calI} \calB_{r_i + r}(\bfp_i(t)) \right) \bigcup \Omega^+.
\end{equation}

\begin{figure}[t]
    \centering
    \includegraphics[width=0.49\linewidth,valign=t]{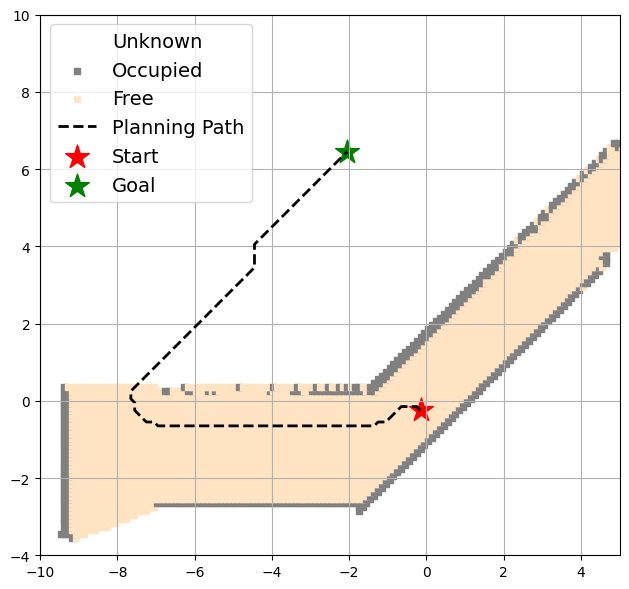}%
    \hfill%
    \includegraphics[width=0.49\linewidth,valign=t]{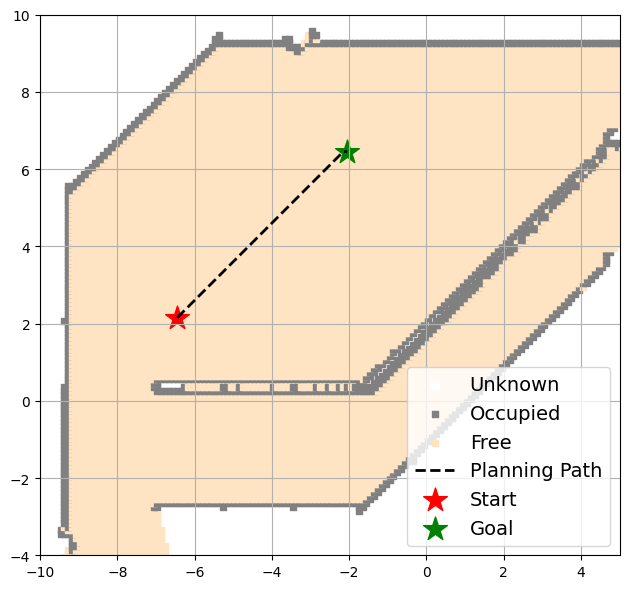}
    \caption{Map update and path re-plan. The left figure shows the partial map and the planned path at the start. The right figure shows an updated map and a re-planned path as the robot proceeds to the goal.}
    \label{fig:map_and_path_update}
\end{figure}

\begin{definition}\label{def:safe}
The robot is \emph{safe} if it remains collision-free with respect to both static and moving obstacles at all times, i.e., $\bfp(t) \in \Cl(\calW \setminus \calO^+(t)$), $\forall t \geq 0$, where $\Cl(\cdot)$ denotes the closure of a set.
\end{definition} 

The robot is equipped with a sensor, such as a LiDAR or depth camera, that provides distance measurements from the robot's position $\bfp(t)$ to the obstacles $\calO^+(t)$. We assume that the robot is able to measure the positions $\bfp_i(t)$ and velocities $\bfv_i(t)$ of the moving obstacles (e.g., using an object tracking algorithm \citep{MOTChallenge20}). 

The quadratic distance between a point $\bfa$ and a set $\calB$ is defined as:
\begin{equation}
    d_{\bfQ}(\bfa, \calB) \coloneqq \inf_{\bfb \in \calB} \| \bfa - \bfb \|_{\bfQ},
\end{equation}
where $\|\bfx\|_{\bfQ} \coloneqq \sqrt{\bfx^\top \bfQ \bfx}$ for a symmetric positive-definite matrix $\bfQ$. Similarly, the quadratic distance between two sets $\calA$, $\calB$ is defined as:
\begin{equation}
    d_{\bfQ}(\calA, \calB) \coloneqq \inf_{\bfa \in \calA, \bfb \in \calB} \norm{\bfa - \bfb}_\bfQ.
\end{equation}
When $\bfQ = \bfI$, $d_\bfQ$ reduces to the Euclidean distance, and we drop the subscript to simplify the notation.

The distance measurements provided by the sensor are used to construct an occupancy map of the environment that contains free space, occupied space, and unknown space. Due to the limited sensing range of the robot, the evolving partial map is used to repeatedly re-plan a path to a desired goal location. As the robot moves, it receives new measurements, updates the map, and re-plans the path, e.g., using a motion planning algorithm such as A* \citep{Astar_hart1968formal} or RRT \citep{RRT} assuming that the unknown space is free, as illustrated in Fig.~\ref{fig:map_and_path_update}.

A \emph{path} is a piecewise continuous function $\rho: \brl{0,1} \rightarrow \calW$ that maps a path-length parameter $\sigma \in \brl{0,1}$ to the interior of the free space considered by the planner. A path starts at $\rho(0)$ and ends at $\rho(1)$. 

Our objective is to design a control policy for the robot with kinematics in \eqref{eq:kinematic_unicycle} to track a planned path $\rho(\sigma)$ while remaining safe at all times (Def.~\ref{def:safe}).

\section{Robot System Design}

\begin{figure}[t]
    \centering
    \includegraphics[width=0.9\linewidth]{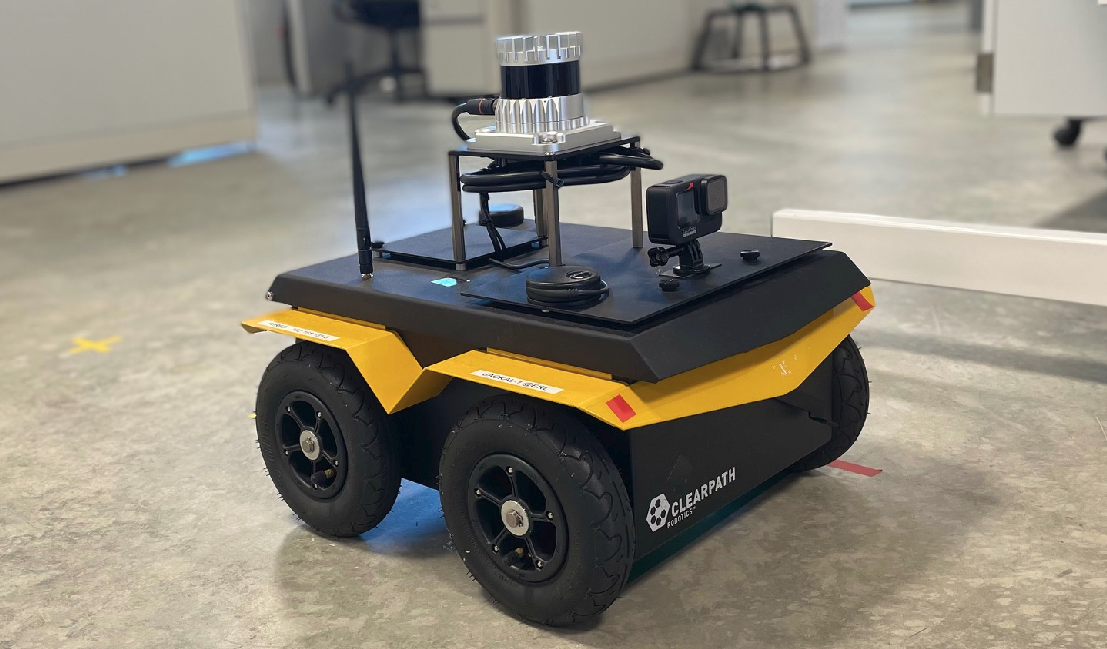}
    \caption{Jackal unmanned ground vehicle.}
    \label{fig:jakcal_hw}
\end{figure}

This section overviews the robot system we used to evaluate our safe autonomous navigation method, including the robot's computation, sensing, and communication hardware and its localization, mapping, motion planning, and control software.

\subsection{Hardware Architecture}

We used a Clearpath Jackal unmanned ground vehicle, shown in Fig.~\ref{fig:jakcal_hw}. The Jackal is a differential-drive robot with four wheels and dimensions $508 \times 430 \times 250$ mm. It weighs approximately $17$ kg and runs at $2.0$ m/sec top speed. 

The robot was equipped with an on-board computer with an Intel i7-9700TE CPU with 32GB RAM, an Ouster OS1-32 LiDAR and a 9-axis IMU UM7. It can be controlled manually by a Bluetooth joystick or remotely by Wi-Fi access. The joystick was used for manual data collection or as an emergency stop controller. A local network, configured through a standard router, was used to monitor the status of our algorithms through visualization tools in the Robot Operating System (ROS) \citep{quigley2009ros}.

\subsection{Software Architecture}
\begin{figure}[t]
    \centering
    \includegraphics[width=\linewidth]{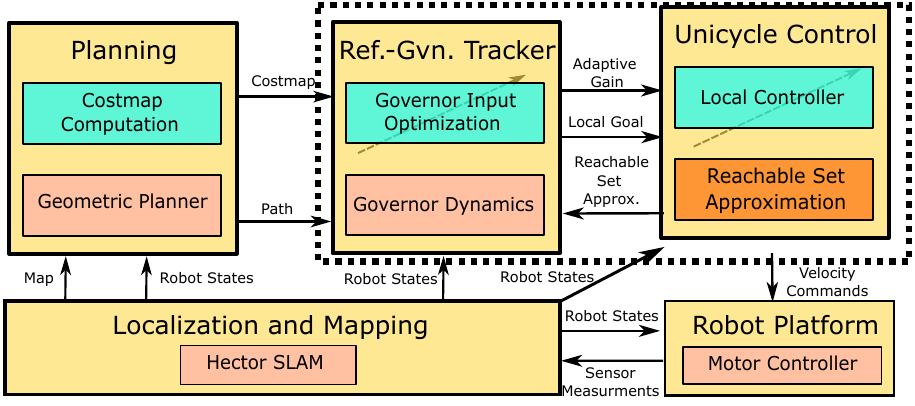}
    \caption{System architecture overview. This paper focuses on the planning block (top left) and the control block (top right, dashed).}
    \label{fig:architecture}
\end{figure}

The software components fall in three categories: localization and mapping, planning, and control. We used ROS to exchange messages among these components and the Gazebo physics simulator \citep{koenig2004design} to test and debug our algorithms before the hardware deployment. Fig.~\ref{fig:architecture} provides an overview of the software architecture.

\textbf{Localization and Mapping.}
We used Hector SLAM \citep{hectorSLAM}, which takes 2D LiDAR scans as input and provides a 2D occupancy grid map and robot poses (positions and orientations) in the map frame as output. The ROS package \verb|pointcloud_to_laserscan| was used to convert point clouds from the Ouster LiDAR to 2D LiDAR scans.

\textbf{Planning.}
The planning module consists of two parts: cost map construction and geometric path planning using the $A^*$ algorithm \citep{Astar_hart1968formal}. A cost map is constructed as a distance field over the occupancy grid map provided by Hector SLAM and is tuned to trade off obstacle clearance with traveled distance. Details about the cost map construction are discussed in Sec.~\ref{sec:costmap_design}.

\textbf{Control.}
We used a low-level velocity controller provided by Clearpath Robotics to convert linear and angular velocity inputs to motor torque inputs. Our main contribution is the design of a trajectory tracking controller to enforce safety constraints with respect to static and dynamic obstacles. Our controller consists of two parts: (1) a stabilizing control law with associated convex reachable set over-approximation and (2) an adaptive trajectory tracker that balances stabilization and safety constraints. The control law is designed to stabilize the differential-drive robot to a desired equilibrium and to provide a convex set that contains the predicted trajectory from the initial robot state to the equilibrium state. To bridge the gap between the geometric path provided by the planner and the point stabilization capability of the controller, we design an environment-aware safe tracker. The tracker generates local reference points along the path for the controller adaptively based on comparisons between the local safe region around the robot and the predicted controlled reachable set. On the one hand, when there are no obstacles around the robot, the reachable set can be large, corresponding to a distant reference point along the path and high robot velocity, without endangering safety. On the other hand, when there are obstacles around, the local safe space is small requiring a small reachable set to ensure safety and, hence, a nearby reference point along the path and low robot velocity.

\section{Safe Tracking via Reference Governor}
\label{sec:safe_tracking}

This section describes our cost map design for path planning (Sec.~\ref{sec:costmap_design}) and our control design for environment-aware safe tracking (Sec.~\ref{sec:ref_gov_tracker}).

\begin{figure*}[t]
	\centering
	\begin{subfigure}{0.27\linewidth}
		\centering
		\includegraphics[width=\linewidth]{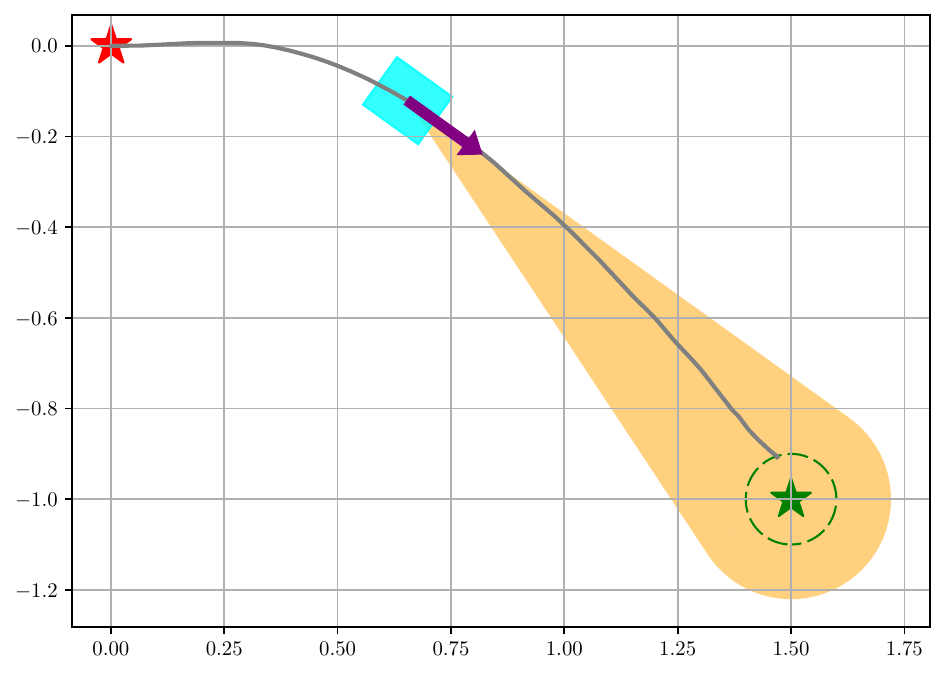}
		\caption{reachable set}
		\label{fig:cone_reacable_set}
	\end{subfigure}
	\begin{subfigure}{0.26\linewidth}
		\centering
		\includegraphics[width=\linewidth]{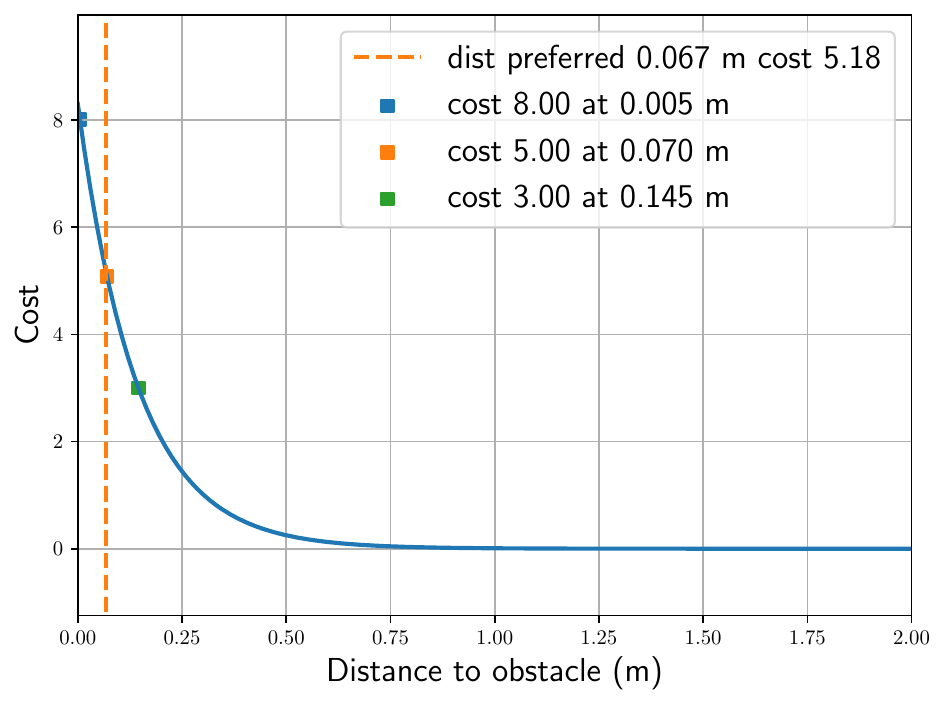}
		\caption{cost curve}
		\label{fig:cost_curve}		
	\end{subfigure}
	\begin{subfigure}{0.23\linewidth}
		\centering
		\includegraphics[width=\linewidth]{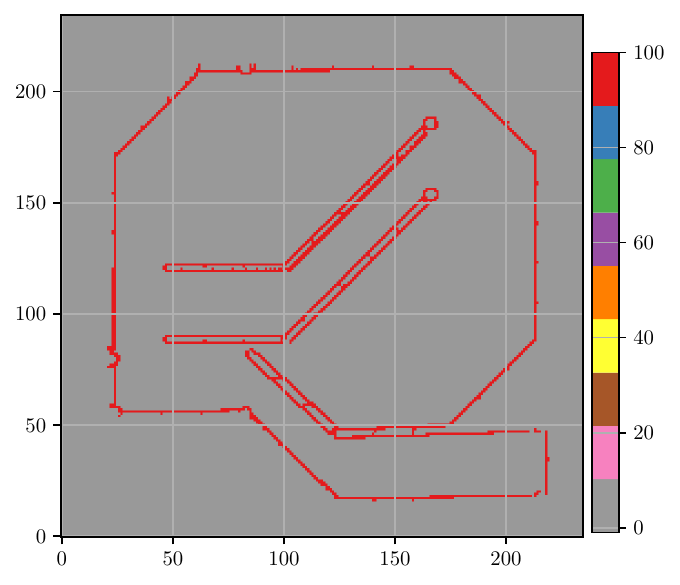}
		\caption{occupancy grid map}
		\label{fig:ogm_jackal_race}		
	\end{subfigure}
	\begin{subfigure}{0.22\linewidth}
		\centering
		\includegraphics[width=\linewidth]{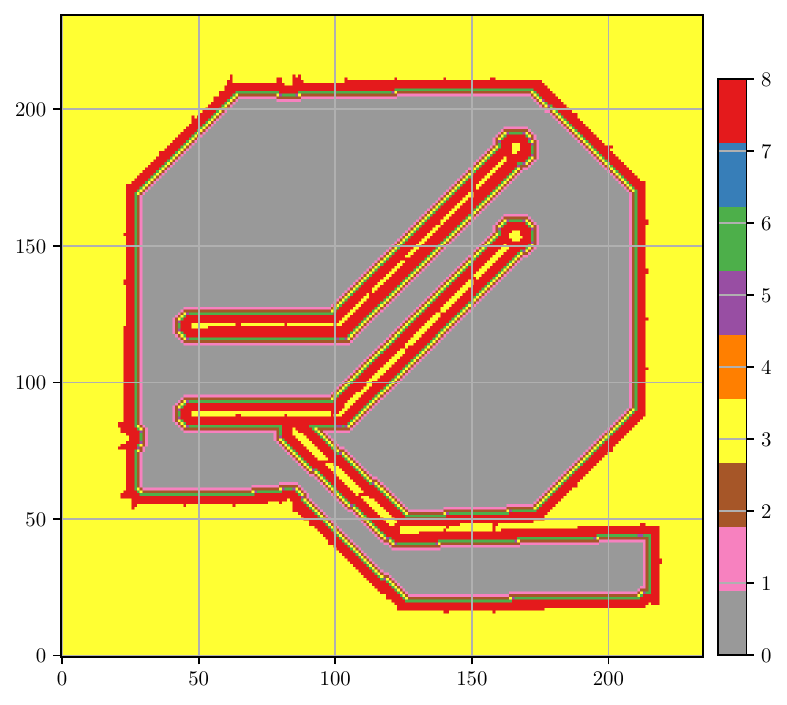}
		\caption{cost map}
		\label{fig:costmap_jackal_race}		
	\end{subfigure}	
	\caption{Ice-cream cone approximation (orange) of the reachable set for a differential-drive system \eqref{eq:kinematic_unicycle} with control law in \eqref{eq:cone_controller} is shown in (a). The start and goal positions are depicted as red and green stars, respectively. The robot is shown as a cyan rectangle with a purple arrow showing its orientation. The obstacle clearance cost design for motion planning is shown in (b). Note that the obstacle clearance \eqref{eq:obstacle_clearance} is a function of $\bfp$, but we plot the value of $c(\bfp)$ versus $d(\bfp, \calO^+)$. An occupancy grid map of a simulated environment is shown in (c), with cell values equal to the probability (in percent) of being occupied by obstacles. The corresponding clearance cost is shown in (d), with clearance cost set to $3$ for unknown regions in the occupancy grid map.}
    \label{fig:jakcal_footprint_costcurve_hectorOGM_costmap}
\end{figure*}

\subsection{Cost Function Design}
\label{sec:costmap_design}

Given the robot position $\bfp(t)$ at time $t$ and an occupancy map approximating the obstacle space $\calO^+(t)$ obtained from SLAM, we periodically re-plan a path $\rho$ to a desired goal position $\bfp^*$. We assume that unknown parts of the map are free for the purpose of motion planning. The motion planning problem at time $t$ is formulated as:
\begin{equation} \label{eq:motion_planning}
\begin{aligned}
    \min_{\rho} & \; C(\rho),\\
    \text{s.t.} &\; \rho(0) = \bfp(t), \; \rho(1) = \bfp^*,\\
    &\; \rho(\sigma) \in \calW \setminus \calO^+(t), \; \forall \sigma \in (0,1),
\end{aligned}
\end{equation}
where the constraints require that the path $\rho$ starts at the current robot position $\bfp(t)$, ends at the goal $\bfp^*$, and the robot remains collision-free along the way with respect to the occupancy information $\calO^+(t)$ at time $t$. We restrict the path $\rho(\sigma)$ to be a piecewise-linear function characterized by $N + 1$ vertices $\calP \coloneqq \{ \rho(\sigma_i) \}_{i = 0}^{N}$ defined as:
\begin{equation} \label{eq:planned_path}
    \rho(\sigma) \coloneqq \rho(\sigma_i) + \frac{\sigma - \sigma_i}{\sigma_{i + 1} - \sigma_i}(\rho(\sigma_{i + 1}) - \rho(\sigma_{i})), 
\end{equation}
for $\sigma_i \leq \sigma < \sigma_{i + 1}$. 

The motion planning problem in \eqref{eq:motion_planning} can be solved by a variety of motion planning algorithms, such as $A^*$ \citep{Astar_hart1968formal}, PRM \citep{PRM} and RRT \citep{RRT}. We focus on designing a cost function $C(\rho)$ to trade off obstacle clearance and travel distance of the form: 
\begin{equation} \label{eq:cost_function}
    C(\rho) \coloneqq \sum_{i = 0}^{N} \| \rho(\sigma_{i + 1}) - \rho(\sigma_i) \| + c(\rho(\sigma_{i + 1})), 
\end{equation}
where the first term captures travel distance and the second term captures obstacle clearance. The \emph{obstacle clearance} term $c : \bbR^2 \rightarrow \bbR$ is defined as:
\begin{equation} \label{eq:obstacle_clearance}
    c(\bfp) \coloneqq c_u \exp(-\kappa d(\bfp, \calO^+)), 
\end{equation}
where $c_u > 0$ is a parameter placing an upper bound on the clearance cost term and $\kappa > 0$ is an exponential decay rate parameter. One way to obtain the distance $d(\bfp, \calO^+)$ in practice is via a distance transform \citep{opencv_library} applied to the occupancy map. Since $c(\bfp)$ in \eqref{eq:cost_function}, \eqref{eq:obstacle_clearance} is positive, the Euclidean distance $h(\bfp) = \norm{\bfp-\bfp^*}$ is an admissible and consistent heuristic \citep{ARAstar} for the motion planning problem in \eqref{eq:motion_planning}.

The safety constraint $\rho(\sigma) \in \calW \setminus \calO^+(t)$ in \eqref{eq:motion_planning} can be enforced by setting a minimum acceptable obstacle clearance $c(\rho(\sigma)) \leq c_f$ and considering positions with obstacle clearance cost larger than $c_f$ occupied. 

\begin{proposition} \label{lemma:reformulated_safety}
    Consider the safety constraint $\rho(\sigma) \in \calW \setminus \calO^+(t)$ in \eqref{eq:motion_planning} and the obstacle clearance cost defined in \eqref{eq:obstacle_clearance}. Let $c_f < c_u$. The safety constraint in \eqref{eq:motion_planning} can be reformulated as:
    \begin{equation} \label{eq:planning_cutoff}
        c(\rho(\sigma)) \leq c_f, \quad \forall \sigma \in (0, 1).
    \end{equation}
\end{proposition}

\begin{proof}
    By the definition of the obstacle clearance cost in \eqref{eq:obstacle_clearance}, $c(\rho(\sigma)) \leq c_f < c_u$ implies $d(\bfp, \calO^+) > 0$. 
\end{proof}

The obstacle clearance cost with $c_u = 8.3$ and $\kappa = 7$ is visualized in Fig.~\ref{fig:cost_curve}. Based on Fig.~\ref{fig:cost_curve}, we set $c_f = 5$ to ensure $\rho(\sigma) \in \calW \setminus \calO^+(t)$. An occupancy grid map and the corresponding obstacle clearance cost are shown in Fig.~\ref{fig:ogm_jackal_race} and Fig.~\ref{fig:costmap_jackal_race}.

\subsection{Reference Governor Safe Tracker}
\label{sec:ref_gov_tracker}

Given a path $\rho(\sigma)$ obtained from the motion planning problem in \eqref{eq:motion_planning}, our next objective is to design a control policy for the differential-drive robot in \eqref{eq:kinematic_unicycle} to track $\rho(\sigma)$ subject to the safety constraint in \eqref{eq:planning_cutoff}. We first present existing results considering the case where no moving obstacles are present in Sec.~\ref{sec: prelim_on_static}. Our main contribution is presented in Sec.~\ref{sec:ref_gvn_extension}, where we extend the existing results to dynamic environments.

\subsubsection{Safe Tracking in Static Environments}
\label{sec: prelim_on_static}
We focus on designing a control law to track a reference path $\rho(\sigma)$ subject to the safety constraint $\bfp(t) \in \calW \setminus \Omega^+$ that the robot remains in the free space of the static environment for all $t$. First, we consider stabilization to a point $\bfp^*$ in the absence of constraints and approximate the reachable set of the robot. We use the control policy proposed by \citet{omur2023feedback}:
\begin{equation}\label{eq:cone_controller}
    \bfk(\bfx,\bfp^*) = \begin{bmatrix}
        v(\bfx,\bfp^*)\\\omega(\bfx,\bfp^*)
    \end{bmatrix} = \begin{bmatrix}
        k_v\; e_v\\ k_\omega\; \arctan\prl{e_v^\perp /  e_v}
    \end{bmatrix},
\end{equation}
where $\bfx = (\bfp, \theta)$ is the robot state, $k_v > 0$ and $k_\omega > 0$ are control gains for the linear and angular velocities, and the error terms $e_v$ and $e_v^\perp$ are defined as:
\begin{equation*}
	e_v = \begin{bmatrix}
	\cos \theta \\
	\sin \theta 
	\end{bmatrix}^\top (\bfp^* - \bfp), \quad
	e_v^\perp =
	\begin{bmatrix}
	-\sin \theta \\
	\cos \theta 
	\end{bmatrix}^\top(\bfp^* - \bfp).
\end{equation*}
We set $\omega(\bfx,\bfp^*) = 0$ when $\bfp = \bfp^*$. 
\begin{proposition}[{\citet[Lemma~1]{omur2023feedback}}] \label{prop:cone_control_converge}
    The control law in \eqref{eq:cone_controller} applied to the system in \eqref{eq:kinematic_unicycle} asymptotically steers all initial states $(\bfp_0, \theta_0)$ in $\bbR^2 \times [-\pi, \pi)$ to a given goal position $\bfp^* \in \bbR^2$, i.e., the closed-loop trajectory $(\bfp(t), \theta(t))$ satisfies $\lim_{t \rightarrow \infty} \bfp(t) = \bfp^*$.
\end{proposition}
Accurate prediction of the closed-loop robot trajectory is important for fast and safe path following. However, the closed-loop system is nonlinear and predicting its trajectory requires numerical integration. An alternative is to compute a reachable set that provides bounds on the possible trajectories. \citet{omur2023feedback} show that trajectories generated by \eqref{eq:kinematic_unicycle} under control law \eqref{eq:cone_controller} are contained in a reachable set with ice-cream cone shape, illustrated in Fig.~\ref{fig:cone_reacable_set}.

\begin{proposition}[{\citet[Proposition~4]{omur2023feedback}}]\label{prop:motion_prediction}
    Consider the set $\calM(\bfx, \bfp^*) \coloneqq \calR \prl{\bfp, \bfp^*, \norm{e_v^\perp}}$, where: 
    \begin{equation*} 
        \calR \prl{\bfa, \bfb, r} \coloneqq \crl{\bfa + \alpha(\bfz - \bfa) \mid \alpha \in \brl{0, 1}, \bfz \in \calB_r(\bfb)}.
    \end{equation*}
    For any given goal position $\bfp^*$ and any initial condition $\bfx_0 =  (\bfp_0, \theta_0)$, the control law \eqref{eq:cone_controller} applied to system \eqref{eq:kinematic_unicycle} renders $\calM(\bfx_0, \bfp^*) \times [-\pi, \pi)$ forward invariant.
\end{proposition}
\begin{figure}[t]
    \centering
    \includegraphics[width=0.9\linewidth]{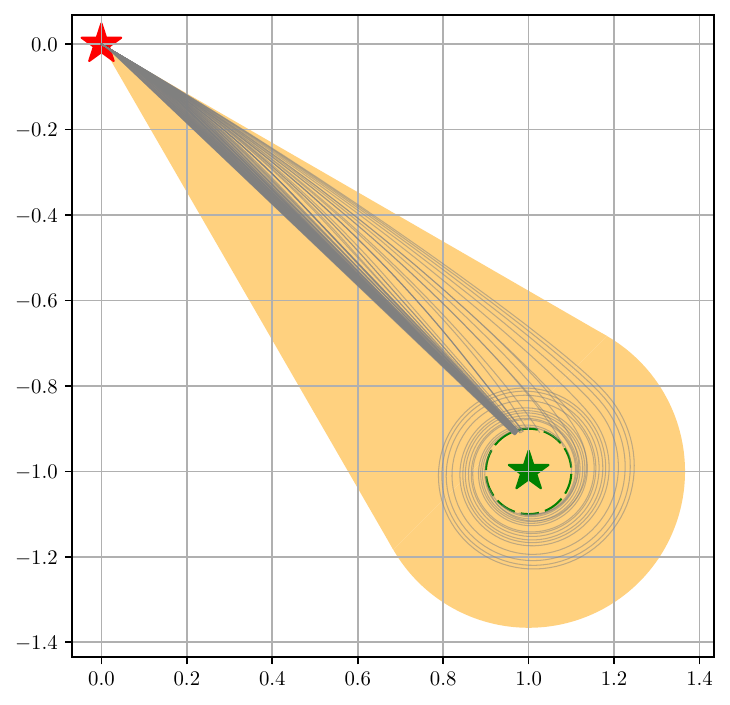}
    \caption{Simulation of differential-drive trajectories (gray curves) under the control law in \eqref{eq:cone_controller} with $k_v = 2.0$, $k_{\omega} = 5.0$, and control bounds $|v| \leq v_{max}$ and $| \omega | \leq \omega_{max}$ such that $v_{max}, \omega_{max} \in \{0.1, 0.2, \ldots, 1.0\}$. The robot position and goal position are depicted as red and green stars, respectively. The reachable set derived in Prop.~\ref{prop:motion_prediction} is shown as a yellow region. To ease visualization, the trajectories are terminated when they enter a ball of radius $0.1$ centered at the goal (green dashed circle). Despite variations in the control limits, all trajectories remain entirely within the reachable set, illustrating that the reachable set depends solely on the robot's position and the governor's state.}
    \label{fig:reachable_set}
\end{figure}

The reachable set approximation $\calM(\bfx, \bfp^*)$ is convex, making distance computations efficient. Additionally, it is more accurate than ellipsoidal reachable set approximations derived from the Lyapunov function associated with \eqref{eq:cone_controller}, resulting in less conservative distance computation. 

As stated in Prop.~\ref{prop:motion_prediction}, the reachable set prediction depends only on the robot position and the goal position, and is independent of the particular control gains $k_v$ and $k_{\omega}$. We illustrate this with a simulation using fixed control gains $k_v = 2.0$ and $k_{\omega} = 5.0$ and varying control bounds $|v| \leq v_{max}$ and $| \omega | \leq \omega_{max}$, with $v_{max}, \omega_{max} \in \{0.1, 0.2, \ldots, 1.0\}$. We generated 100 differential-drive trajectories using different combinations of the control bounds. The results are shown in Fig.~\ref{fig:reachable_set}. We observe that, for a fixed linear velocity bound, as the angular velocity bound approaches zero ($\omega_{max} \to 0$), the trajectory lies closer to the boundary of the predicted reachable set. In contrast, for larger angular velocity bounds, the trajectory tends to approximate a straight line connecting the starting point to the goal point.

Given the controller in \eqref{eq:cone_controller} and the reachable set approximation in Prop.~\ref{prop:motion_prediction}, we enforce the safety constraint $\bfp(t) \in \calW \setminus \Omega^+$ using the reference governor method \citep{RG_Omur_ICRA17, RG_nicotra2018_ERG}. A \emph{reference governor} is a first-order virtual system with dynamics:
\begin{equation} \label{eq:governor_dynamics}
\dot{\bfg} = -k_g(\bfg - \bfu_g), 
\end{equation}
where $\bfg \in \bbR^2$ is the governor state, $\bfu_g \in \bbR^2$ is the governor input, and $k_g > 0$ is a parameter.
A governor is used to decouple the stabilization of a dynamical system from the enforcement of safety constraints. 

We use the governor state $\bfg$ as the reference point in the stabilizing controller $\bfk(\bfx,\bfg)$ in \eqref{eq:cone_controller}. This way, the closed-loop system aims to reach the governor state, while the governor can move along the reference path $\rho(\sigma)$ in a way to ensure that the reachable set $\calM(\bfx, \bfg)$ remains in free space. To satisfy the safety constraint, the input $\bfu_g$ of the governor system \eqref{eq:governor_dynamics} must be chosen by considering the size of the free space $\Omega^+$ in comparison to the size of the reachable set approximation $\calM(\bfx, \bfg)$. We measure the difference in the sizes of these sets by a distance metric $d(\calM(\bfx, \bfg), \Omega^+)$ and use it to define a set of feasible governor inputs, termed local safe zone.

\begin{definition}\label{def:LSZ}
A \emph{local safe zone} is a set, determined by the joint system-governor state $(\bfx, \bfg)$ and the distance $d(\calM(\bfx, \bfg), \Omega^+)$ between the reachable set $\calM(\bfx, \bfg)$ and the static obstacle space $\Omega^+$: 
\begin{equation*} 
    \LS(\bfx,\bfg) \coloneqq \crl{ \bfq \in \bbR^2 \mid \norm{\bfq - \bfg}^2 \leq d(\calM(\bfx, \bfg), \Omega^+)}.
\end{equation*}
\end{definition}

\begin{figure}[t]
\centering
\includegraphics[width=\linewidth]{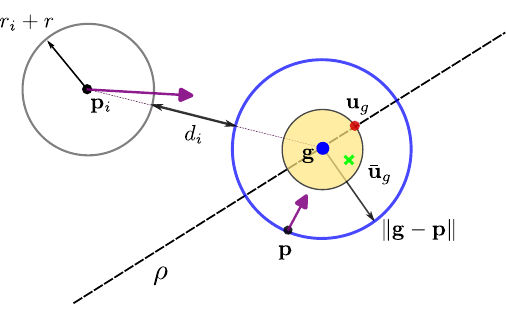}	
\caption{Geometric relationship between the local safe zone $\LS$ (yellow ball), governor input $\bfu_g$ (red dot) and modified governor input $\bar{\bfu}_g$ (green cross). The blue ball $\calB_{\norm{\bfg-\bfp}}(\bfg)$ represents an outer approximation of the reachable set $\calM(\bfx, \bfg)$, the gray ball $\calB_{r_i + r}(\bfp_i)$ represents the inflated obstacle space for moving obstacle $i$, and $d_i$ is the distance between the two balls.}
\label{fig:gvn_components}
\end{figure}

The definitions of the local safe zone and governor input are illustrated in Fig.~\ref{fig:gvn_components}. The governor input at system-governor state $(\bfx, \bfg)$ is chosen as the point $\bfu_g$ in the local safe zone $\LS(\bfx,\bfg)$ that is furthest along the reference path $\rho$:
\begin{equation} \label{eq:lpg}
\bfu_g = \rho(\bar{\sigma}),  \;\; \bar{\sigma} = \arg\!\max_{\sigma \in [0,1]}\! \crl{ \sigma \mid  \rho(\sigma) \in \LS(\bfx,\bfg)}\!.
\end{equation}

Combining the system in \eqref{eq:kinematic_unicycle} with controller $\bfk(\bfx, \bfg)$ in \eqref{eq:cone_controller} and the governor dynamics \eqref{eq:governor_dynamics} with controller in \eqref{eq:lpg}, we have the closed-loop system-governor dynamics:
\begin{equation} \label{eq:robot_governor_dynamics}
\begin{bmatrix}
\dot{\bfx} \\
\dot{\bfg}
\end{bmatrix} = \begin{bmatrix}
G(\bfx) \bfk(\bfx, \bfg) \\
-k_g \bfg
\end{bmatrix} + \begin{bmatrix}
\mathbf{0} \\
k_g I
\end{bmatrix} \bfu_g = A(\bfx,\bfg) + B\bfu_g.
\end{equation}
The closed-loop system-governor dynamics satisfy the safety constraint $\bfp(t) \subset \calW \setminus \Omega^+$.

\begin{proposition} \label{thm:main_result}
Given a reference path $\rho$, consider the closed-loop system-governor dynamics in \eqref{eq:robot_governor_dynamics}. Suppose that the initial state $(\bfx_0, \bfg_0)$ satisfies:
%
\begin{equation} \label{eq:thm_ic}
    d(\calM(\bfx_0, \bfg_0), \Omega^+) > 0,\quad  \bfg_0 = \rho(0) \in \calW \setminus \Omega^+.
\end{equation}
Then, the position $\bfp(t)$ converges to the end of the path $\rho(1)$ without collision, i.e., $\bfp(t) \in \calW \setminus \Omega^+$, $\forall t \geq 0$.
\end{proposition}
\begin{proof}
    The proof follows from the results in~\citet{omur2023feedback, arslan2022time, Li_SafeControl_ICRA20} and the references therein. It consists of two parts. First, we show that the closed-loop dynamics in \eqref{eq:robot_governor_dynamics} ensure the safety of the system position, i.e., $\bfp(t) \in \calW \setminus \Omega^+$, $\forall t \geq 0$. Second, we establish that the system position asymptotically converges to the endpoint of the reference path. \\
    Since the distance between the reachable set and the static obstacle space $d(\calM(\bfx, \bfg), \Omega^+)$ is Lipschitz continuous \citep[Proposition~9]{omur2023feedback}, one can consider a partition of the time interval $[0, \infty)$ based on distinct time instances $(t_0=0, t_1, ..., t_i, ...)$ with $t_i < t_{i+1}$ such that $d(\calM(\bfx(t), \bfg(t)), \Omega^+)$ is either positive or zero over $[t_i, t_{i+1})$, and alternates between these cases across the partition. Given such a time interval $[t_i, t_{i+1})$, we distinguish two cases.
    \begin{itemize}
        \item If $d(\calM(\bfx(t), \bfg(t)), \Omega^+) > 0$ for any $t \in [t_i, t_{i+1})$, then both the reachable set and the robot position are in the interior of the free space, i.e., $\calM(\bfx(t), \bfg(t)) \subset \Int(\calW \setminus \Omega^+)$ and $\bfp(t) \in \calW \setminus \Omega^+$, because, by definition $\bfp(t) \in \calM(\bfx(t), \bfg(t))$.
        
        \item If $d(\calM(\bfx(t), \bfg(t)), \Omega^+) = 0$ for any $t \in [t_i, t_{i+1})$, we have $\Int(\calM(\bfx(t), \bfg(t))) \bigcap \Int(\Omega^+) = \emptyset$. To show this, notice that the distance between the reachable set and the static obstacle space $d(\calM(\bfx, \bfg), \Omega^+)$ is Lipschitz continuous and was strictly positive in the previous time interval, i.e., $d(\calM(\bfx(t), \bfg(t)), \Omega^+) > 0$ for any $t \in [t_{i-1}, t_{i})$. As a result, at time $t=t_i$, the reachable set and the obstacle space are in contact at their boundaries without overlapping interiors: $\Int(\calM(\bfx, \bfg)) \bigcap \Int(\Omega^+) = \emptyset$, and $\partial \calM(\bfx, \bfg) \bigcap \partial \Omega^+ \neq \emptyset$, where $\partial(\cdot)$ denotes the boundary of a set. Hence, it follows from $\dot{\bfg} = \mathbf{0}$ over $[t_i, t_{i+1})$ and Prop.~\ref{prop:motion_prediction} that under the proposed control policy \eqref{eq:cone_controller}, the robot position trajectory satisfies $\bfp(t) \in \calM(\bfx(t), \bfg(t)) \subset \Cl(\calW \setminus \Omega^+)$ for all $t \in [t_i, t_{i+1})$.
    \end{itemize}
    Thus, the position of the system is collision-free under the closed-loop dynamics \eqref{eq:robot_governor_dynamics}.

    To establish convergence to the goal position, note that $\dot{\bfg} = \mathbf{0}$ if and only if $\bfg = \rho(1)$ or $d(\calM(\bfx, \bfg), \Omega^+) = 0$. To conclude that $\rho(1)$ is the only stable governor state under governor dynamics \eqref{eq:governor_dynamics}, observe that condition $d(\calM(\bfx, \bfg), \Omega^+)$ might be zero only for a finite time. Since the distance between the system position and the governor state $\| \bfp - \bfg \|$ is Lipschitz continuous and asymptotically decays to zero under control policy \eqref{eq:cone_controller}, as shown in Prop.~\ref{prop:cone_control_converge}. Moreover, the reference path lies in the interior of the free space, i.e., $d(\rho(\omega), \Omega^+) > 0$ for all $\omega \in [0, 1]$. Hence, the governor dynamics might be zero only for a finite duration away from $\rho(1)$, and always stays strictly nonzero for at least some finite time. Thus, since the control policy \eqref{eq:cone_controller} is point stabilizing, it follows from LaSalle's invariance principle that both the governor state and the system position asymptotically converge to the end of the reference path, which completes the proof of convergence. 
\end{proof}

To make the robot system converge to the goal efficiently, the gain parameter $k_v$ in \eqref{eq:cone_controller} should be well chosen. Based on results from our previous work \citep{Li_SafeControl_ICRA20}, we develop an adaptive gain that considers the heading of the robot with respect to the surrounding environment. We define a directional distance metric that penalizes obstacles along the robot's heading direction $\bfv = \brl{\cos \theta, \sin \theta}^\top$ as the quadratic distance between the reachable set and the obstacle space $d_{\bfQ\brl{\bfv}}(\calM(\bfx, \bfg), \Omega^+)$ where 
\begin{equation} \label{eq:directional_matrix}
    \bfQ \brl{\bfv} = q_2 \bfI + (q_1 - q_2) \frac{\bfv \bfv^\top}{\norm{\bfv}^2},
\end{equation} 
with scalars $q_2 > q_1 > 0$. We obtain an adaptive control gain by computing the ratio of the directional distance metric to the regular one:
\begin{equation} \label{eq:sddm_boost_gain}
    k_v \coloneqq \begin{cases}
        \frac{d_{\bfQ\brl{\bfv}}(\calM(\bfx, \bfg), \Omega^+)}{d(\calM(\bfx, \bfg), \Omega^+)}, &d(\calM(\bfx, \bfg), \Omega^+) > 0, \\
        1, &\text{otherwise}.
    \end{cases}
\end{equation}
\begin{remark}
The gain $k_v$ in \eqref{eq:sddm_boost_gain} is uniformly bounded. Following the proof by \citet{omur2023feedback}, one can verify that the reachable set prediction in Prop.~\ref{prop:motion_prediction} does not change if $k_v > 0$. As a result, Thm.~\ref{thm:main_result} still holds, indicating that this choice of gain does not endanger safety.
\end{remark}
The reference governor design guides the robot to navigate safely and efficiently in static environments. However, when moving obstacles are present, choosing the governor input $\bfu_g$ along the reference path according to \eqref{eq:lpg} is not sufficient to guarantee safety with respect to the moving obstacles. In this case, we need to modify the governor input $\bfu_g$, considering both static and moving obstacles. This motivates our extension of the reference governor formulation to dynamic environments.

\subsubsection{Safe Tracking in Dynamic Environments}
\label{sec:ref_gvn_extension}

When moving obstacles are present, the robot may need to deviate from its planned path to ensure safety. To achieve this, we choose an appropriate input $\bfu_g$ for the governor in \eqref{eq:governor_dynamics}, which guides the robot to execute avoidance maneuvers.

Inspired by safe control synthesis techniques using CBFs \citep{CBF_ames2019ECC}, we formulate a convex optimization problem to modify the governor input $\bfu_g$ in a minimally invasive way to avoid moving obstacles. We define a CBF that encodes the safety of the system-governor system with respect to the $i$th moving obstacles. Consider a ball $\calB_{\norm{\bfp - \bfg}}(\bfg)$ centered at the governor that contains the robot position $\bfp$ and a ball $\calB_{r_i + r}(\bfp_i)$ centered at the $i$th obstacle position $\bfp_i$ with radius $r_i$ inflated by the robot radius $r$. According to \citet[Prop.~8]{omur2023feedback}, the reachable set $\calM(\bfx, \bfg)$ is a subset of the ball $\calB_{\norm{\bfp-\bfg}}(\bfg)$. Hence, the distance between the reachable set $\calM(\bfx, \bfg)$ and the inflated $i$th moving obstacle is lower bounded by the distance between the two balls, which can be computed as the distance between their centers $\|\bfg - \bfp_i\|$ minus the radii:
\begin{equation*}
	d_{i} = \norm{\bfg - \bfp_i} - \norm{\bfg - \bfp} - r_i - r.
\end{equation*}
This is illustrated in Fig.~\ref{fig:gvn_components}. Due to non-differentiability of $d_{i}$ at $\norm{\bfg - \bfp} = 0$, the distance $d_{i}$ can not be used as a CBF directly. However, note that for $a \geq 0$ and $b > 0$, $(a - b)(a + b) = a^2 - b^2 \geq 0$ implies $a - b \geq 0$ since $a + b > 0$. Hence, a candidate CBF can be constructed by setting $a = \norm{\bfg - \bfp_i}$ and $b = \norm{\bfg - \bfp} + r_i + r$:
\begin{equation}
	h_i(\bfx, \bfg, \bfp_i) = \norm{\bfg - \bfp_i}^2 - \prl{ r_i + r + \norm{\bfg - \bfp}}^2.
\end{equation}
Thus, a CBF constraint for moving obstacle $i$ can be formulated as:
\begin{equation} \label{eq:cbf_def}
	s_i(\bfu_g) \coloneqq \dot{h}_i(\bfx, \bfg, \bfp_i, \bfu_g) + \alpha_i(h_i(\bfx, \bfg, \bfp_i)) \geq 0,
\end{equation}
where $\alpha_i(\cdot)$ is a class-$\calK$ function to be designed and the time derivative of the CBF is:
\begin{equation*}
\dot{h}_i(\bfx, \bfg, \bfp_i, \bfu_g) = \begin{bmatrix} \frac{\partial h_i}{\partial \bfx} & \frac{\partial h_i}{\partial \bfg}\end{bmatrix} (A(\bfx,\bfg) + B\bfu_g) + \frac{\partial h_i}{\partial \bfp_i} \bfv_i
\end{equation*}
with $A(\bfx,\bfg)$ and $B$ defined in \eqref{eq:robot_governor_dynamics}.

The CBF constructed above does not take static obstacles into account. According to our discussion in the previous subsection, a governor input that satisfies static obstacle constraints has to lie in the local safe zone (Def.~\ref{def:LSZ}). Thus, we formulate an optimization problem to find a modified governor input $\bar{\bfu}_g$ that deviates minimally from the desired governor input $\bfu_g$ in \eqref{eq:lpg} and \emph{both} lies within the local safe zone, $\bar{\bfu}_g \in \LS(\bfx, \bfg)$, \emph{and} satisfies the CBF constraints, $s_i(\bar{\bfu}_g) \geq 0$:
\begin{equation}
    \label{eq:opt_active_gov}
    \begin{aligned}
        \min_{\bar{\bfu}_g \in \bbR^2} \;\;&\norm{\bar{\bfu}_g - \bfu_g}^2 \\
        \text{s.t.} \quad  s_i(\bar{\bfu}_g) &\geq 0, \;\;\forall i \in \calI  \\
        \bar{\bfu}_g &\in \LS(\bfx, \bfg).
    \end{aligned}
\end{equation}
Note that the functions $s_i$ in \eqref{eq:cbf_def} are linear in $\bar{\bfu}_g$, while the last constraint is quadratic in $\bar{\bfu}_g$. Hence, \eqref{eq:opt_active_gov} is a convex quadratically constrained quadratic program (QCQP), which can be solved in polynomial time \citep{park2017general}.

When moving obstacles endanger the robot's motion, the modified governor input $\bar{\bfu}_g$ guides the governor and subsequently the robot to deviate from the path $\rho(\sigma)$ to reduce collision risk. An illustrative example is shown in Fig.~\ref{fig:gvn_components}, where a modified governor input $\bar{\bfu}_g$ is found to avoid an incoming obstacle.  

\begin{figure}[t]
    \centering
    \includegraphics[width=0.98\linewidth]{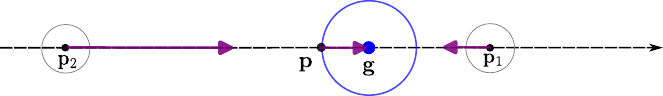}
    \caption{Illustration of the case where the optimization problem \eqref{eq:opt_active_gov} becomes infeasible in the presence of two moving obstacles. The moving obstacles are shown as gray circles centered at $\bfp_1$ and $\bfp_2$, respectively. The robot position is denoted by $\bfp$, and the governor position by $\bfg$. Purple arrows indicate the velocity vectors of the robot and the moving obstacles, reflecting both their directions and magnitudes.}
    \label{fig:cbf_vio}
\end{figure}

According to Thm.~\ref{thm:main_result}, the proposed control synthesis method is guaranteed to be safe in a static environment. When moving obstacles are present, ensuring safety depends on both the robot's motion and the obstacles' motion, and the local safe zone should remain non-empty $\LS(\bfx, \bfg)$. 
According to the local safe zone definition (Def.~\ref{def:LSZ}), the modified governor input can be decomposed as $\bar{\bfu}_g = \bfg + \bfe$, where $\bfe \in \calB_{d(\calM, \Omega^+)}(\bf0)$. The governor dynamics can therefore be expressed as $\dot{\bfg} = k_g \bfe$ and the CBF constraint for moving obstacle $i$ becomes:
\begin{equation}
    \frac{\partial h_i}{\partial \bfg} \bfe + \frac{\partial h_i}{\partial \bfx} \dot{\bfx} + \frac{\partial h_i}{\partial \bfp_i} \bfv_i + \alpha_i(h_i) \geq 0. 
\end{equation}
\cite{breeden2023comp} show that if the CBFs are non-interfering, i.e., $\nabla_{\bfg}^{\top} h_i \nabla_{\bfg} h_j \geq 0$ for all $i, j \in \calI$, then the optimization problem \eqref{eq:opt_active_gov} remains feasible provided that, for each $i \in \calI$:
\begin{equation} \label{eq:cbf_feasible_cond}
    - \| \nabla_{\bfg} h_i \| \leq \sqrt{2} \frac{\frac{\partial h_i}{\partial \bfx} \dot{\bfx} + \frac{\partial h_i}{\partial \bfp_i} \bfv_i + \alpha_i(h_i)}{k_g d(\calM, \Omega^+)}.
\end{equation}
The non-interfering condition on the CBFs precludes cases where the CBF gradients with respect to $\bfg$ point in opposite directions. In particular, the CBF constraints in \eqref{eq:opt_active_gov} can be stacked into a linear system of inequalities: $A \bar{\bfu}_g \leq \bfb$. According to Farkas' Lemma \cite{bertsimas1997introduction}, infeasibility occurs if there exists a vector $\bfy \geq \mathbf{0}$ such that $\bfy^{\top} A = 0$ and $\bfy^{\top} \bfb < 0$, regardless of the size of the local safe zone, as illustrated in Fig.~\ref{fig:cbf_vio}. In the case of two moving obstacles, this condition implies that the CBF gradients with respect to $\bfg$ must point in opposite directions for infeasibility to occur. 

The condition in \eqref{eq:cbf_feasible_cond} further characterizes the size of the local safe zone required for the robot to have sufficient leeway to avoid the moving obstacles, given the configuration of the robot and the moving obstacles. If the local safe zone is too small, e.g., when the robot is close to static obstacles, it becomes difficult to avoid moving obstacles. Conversely, in a relatively empty environment, the condition in \eqref{eq:cbf_feasible_cond} is satisfied as long as $| \nabla_{\bfg} h_i | \neq 0$, which corresponds to the case where the robot position coincides with the governor position, reducing the reachable set to a single point. In this case, any movement of the governor enlarges the reachable set and may therefore compromise the robot’s safety.

\subsection{Generalization to Other Systems}
\label{sec:generalization}

Assuming the existence of a point-stabilizing controller and associated bounded forward-invariant reachable set prediction, our adaptive reference governor tracking method can be generalized to other nonlinear systems. Consider a general nonlinear system:
\begin{equation} \label{eq:generalized_system}
    \dot{\bfx} = \bff(\bfx, \bfu), 
\end{equation}
with state $\bfx = (\bfp, \bfs)$, including position $\bfp \in \bbR^{n_{\bfp}}$ and other variables $\bfs \in \bbR^{n_{\bfs}}$, and input $\bfu \in \bbR^{n_{\bfu}}$. We generalize the workspace $\calW \subset \bbR^{n_{\bfp}}$, the static obstacle space $\Omega^+ \subset \calW$, and the obstacle space $\calO^+ \subset \calW$ to the $n_{\bfp}$-dimensional setting, consistent with the definitions in Sec.~\ref{sec:problem_formulation}. The reference path $\rho: [0, 1] \to \calW$ is obtained by solving \eqref{eq:motion_planning}.

A crucial component in our adaptive reference governor formulation is a point-stabilizing control policy. Since deriving such a policy requires knowledge of the system dynamics, we assume its existence.

\begin{assumption} \label{assum:generalized_control_policy}
    There exists a control policy $\bfk(\bfx, \bfp^*)$ such that, when applied to \eqref{eq:generalized_system}, the closed-loop trajectory $(\bfp(t), \bfs(t))$ converges asymptotically from any initial state $(\bfp_0, \bfs_0) \in \bbR^{n_{\bfp} + n_{\bfs}}$ to the desired goal position $\bfp^* \in \bbR^{n_{\bfp}}$. That is, $\lim_{t \to \infty} \bfp(t) = \bfp^*$.  
\end{assumption}

We also assume the existence of a reachable set prediction, which contains all possible closed-loop trajectories under the policy $\bfk(\bfx, \bfp^*)$.

\begin{assumption} \label{assum:generalized_reachable_set}
    There exists a reachable set $\calM(\bfx, \bfp^*)$ such that, for any goal position $\bfp^*$ and initial condition $\bfx_0 = (\bfp_0, \bfs_0)$, the control policy $k(\bfx, \bfp^*)$ applied to system \eqref{eq:generalized_system} renders $\calM(\bfx, \bfp^*) \times \bbR^{n_{\bfs}}$ forward invariant, i.e., $\bfp(t) \in \calM(\bfx(t), \bfp^*)$ for all $t \geq 0$. Moreover, the reachable set is bounded as $\calM(\bfx(t), \bfp^*) \subseteq \calB_{\| \bfp(t) - \bfp^* \|}(\bfp^*)$ for all $t \geq 0$. 
\end{assumption}

Given Assumption~\ref{assum:generalized_control_policy} and Assumption~\ref{assum:generalized_reachable_set}, the reference governor system \eqref{eq:governor_dynamics} can be generalized to guide the nonlinear system in \eqref{eq:generalized_system}. The governor input $\bfu_g$ is specified by \eqref{eq:lpg}, where the local safe zone $\LS(\bfx,\bfg)$ is defined according to Def.~\ref{def:LSZ}. Under these conditions, Prop.~\ref{thm:main_result} remains valid, and the construction of the adaptive governor input $\bar{\bfu}_g$ follows directly from the exposition in Sec.~\ref{sec:ref_gvn_extension}.

\section{Evaluation}
\label{sec:evaluation}

This section evaluates our Environment-Aware Safe Tracking (EAST) method for safe autonomous robot navigation. We begin by testing EAST in simulated static environments. This includes specifying the design parameters in Sec.~\ref{sec:parameters}, analyzing the impact of the obstacle clearance term \eqref{eq:obstacle_clearance} in Sec.~\ref{sec:clearance_cost}, evaluating the adaptive control gain \eqref{eq:sddm_boost_gain} in Sec.~\ref{sec:controlgain}, and comparing EAST with the EVA-Planner \citep{eva_planner2021} in Sec.~\ref{sec:eva_planner}. Next, we evaluate EAST in a large cluttered real-world static environment in Sec.~\ref{sec:static}. Finally, we evaluate EAST in simulated and real-world dynamic environments in Sec.~\ref{sec:dynamic}.

\subsection{Experiment Setup and Parameters.}
\label{sec:parameters}

All experiments share the same parameters and control gains unless explicitly stated otherwise. We used Hector SLAM \citep{hectorSLAM} to obtain robot pose estimates at $20$ Hz and an occupancy grid map with $0.1$ m resolution. The control gain parameters are summarized in Tab.~\ref{tab:ctrl_params}. The class-$\calK$ function in \eqref{eq:cbf_def} is defined as:
\begin{equation}
\alpha_i(h_i) = \gamma_i h_i^2,
\end{equation}
where $\gamma_i$ is a known parameter specified in Tab.~\ref{tab:ctrl_params}.

\begin{table}[t]
    \centering
    \caption{Control gain parameters.}
    \label{tab:ctrl_params}
    \begin{tabular}{ |l|c|c|c|c|c|c| } 
        \hline 
        Parameter 	&$k_g$ & $k_v$ & $k_\omega$ & $\gamma_i$  &	$q_1$	& $q_2$ \\ 
        \hline
        Value 		&$2.0$ & Eq. \eqref{eq:sddm_boost_gain} & $1.5$	&$0.2$	 &$1$		& $9$  \\ 
        \hline
    \end{tabular}
\end{table}

\subsection{Obstacle Clearance Cost Study}
\label{sec:clearance_cost}

\begin{figure*}[t]
    \centering
    \includegraphics[width=0.32\linewidth]{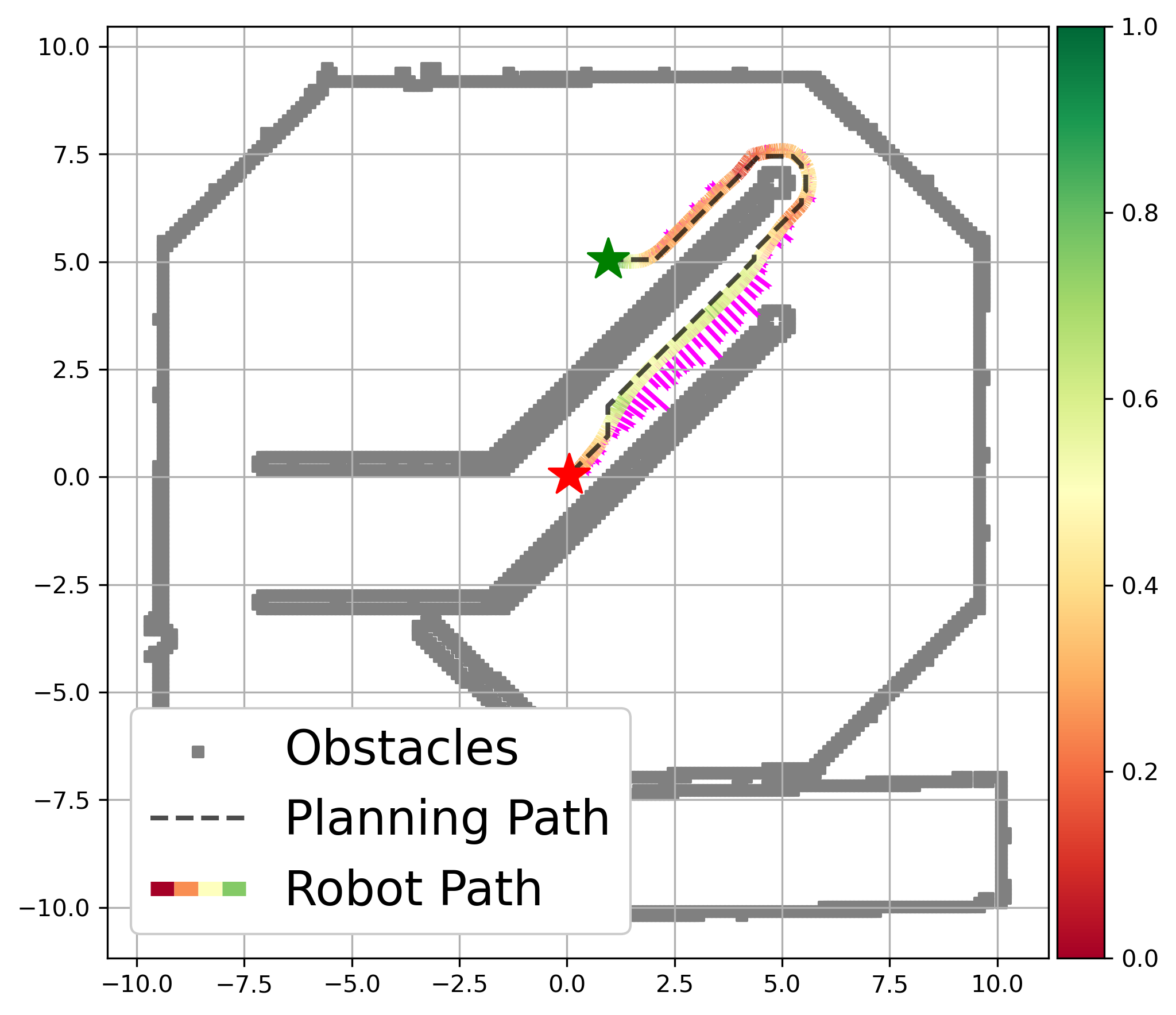}
    \includegraphics[width=0.32\linewidth]{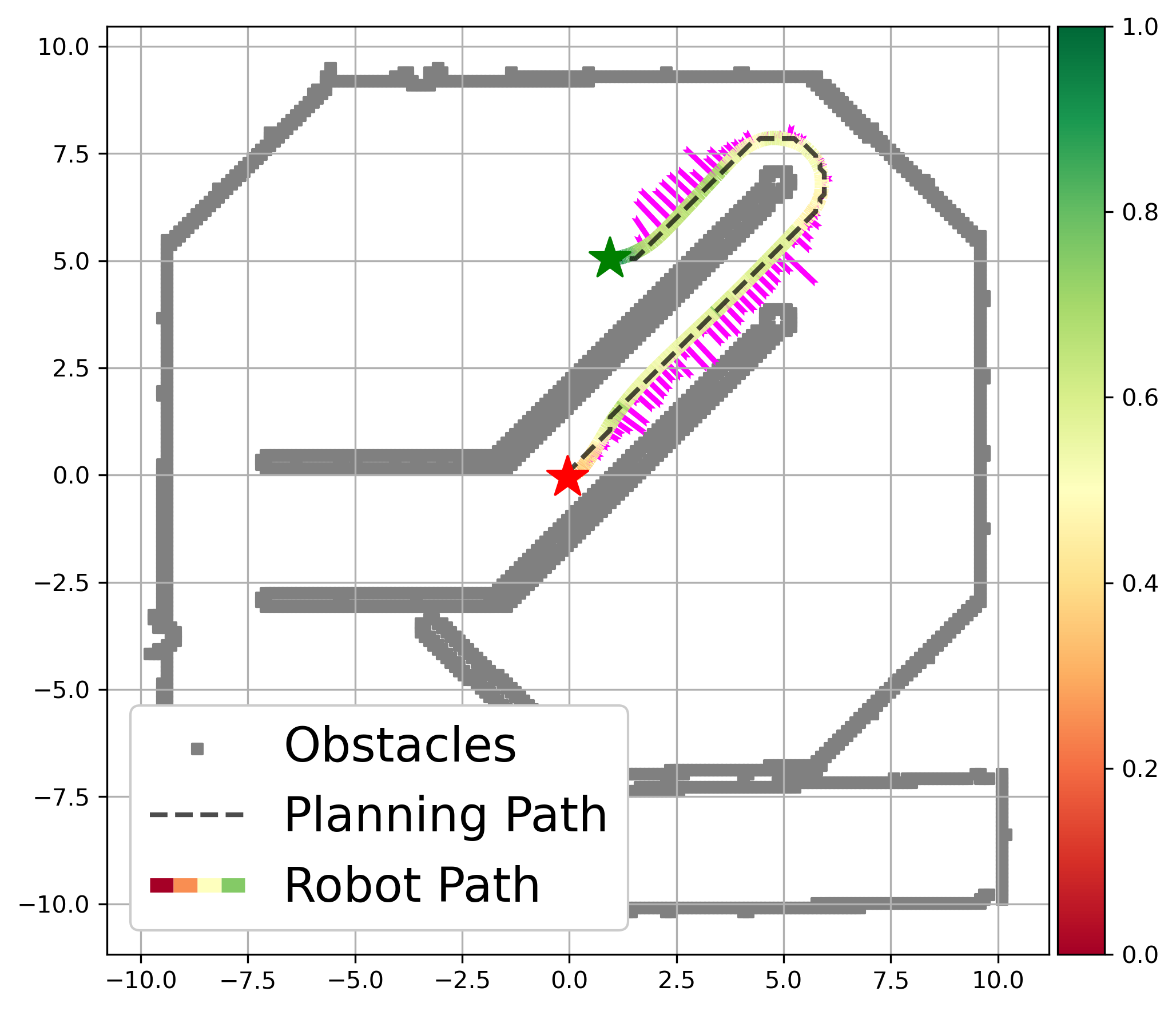}
    \includegraphics[width=0.32\linewidth]{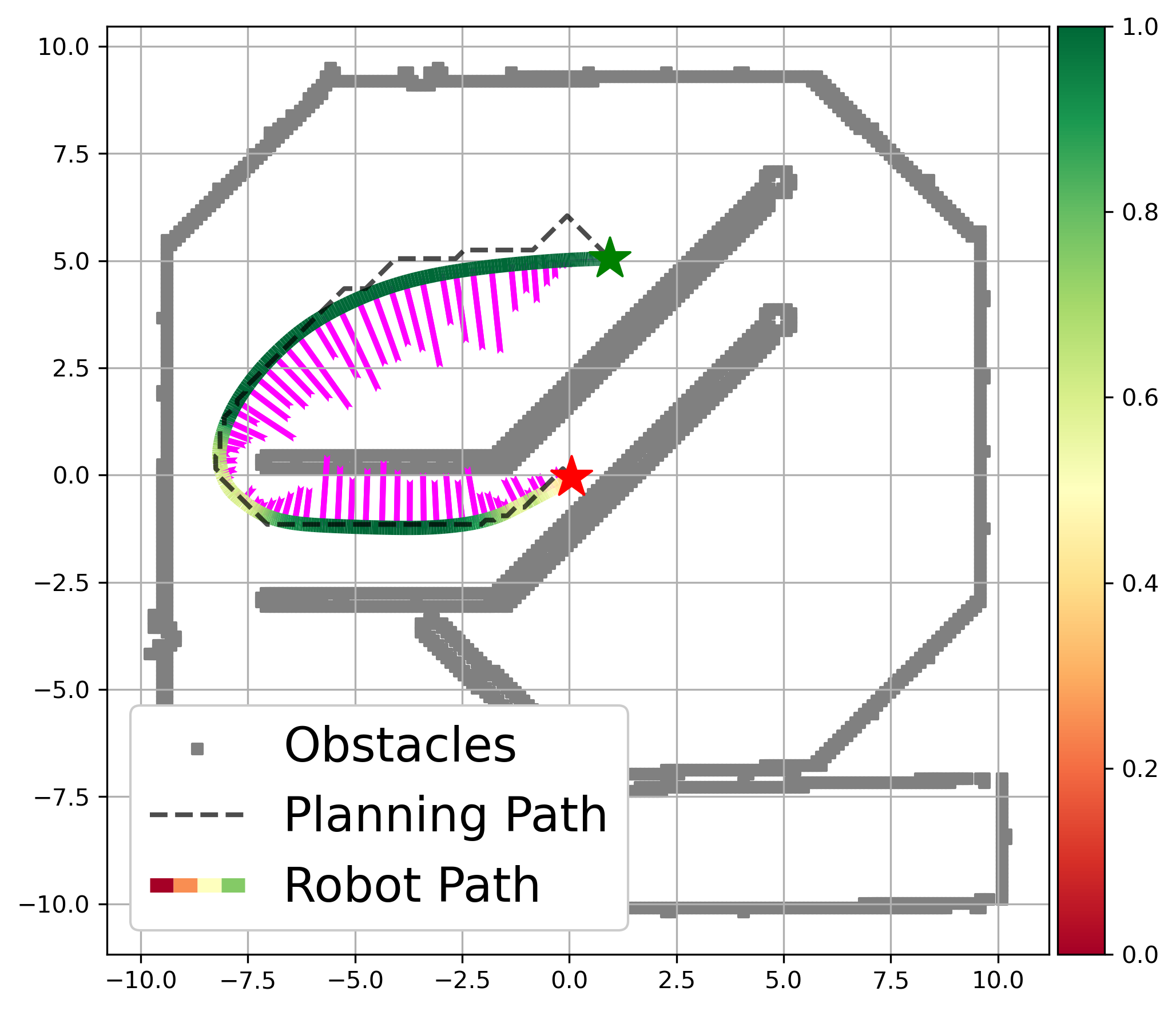}
    \caption{Gazebo simulation with three different obstacle clearance cost designs. The left, middle, and right figures show the results for the minimum, median, and maximum clearance designs, respectively. The paths are shown as a black dashed lines with start and goal denoted by a red and green star, respectively. The colored curve represents the actual robot trajectory, with the robot to the inflated obstacle space distance $d(\bfp, \Omega^+)$ shown by gradient color corresponding to the side color bar. The tracking velocity profiles are shown as magenta arrows perpendicular to the robot's trajectory.}
    \label{fig:cost_curve_design_jackal_env}
\end{figure*}

We study the performance of EAST under different parameters for the obstacle clearance cost in \eqref{eq:obstacle_clearance}, presented in Tab.~\ref{tab:costmap_design}. Paths were planned in a simulated environment for each of the clearance cost designs, shown in Fig.~\ref{fig:cost_curve_design_jackal_env}. EAST can drive the robot to the goal safely in each case. Associated quantitative results are summarized in Tab.~\ref{tab:jackal_race_result}. The results show that, even though the minimum clearance cost design leads to the shortest path, the tracking time is significantly longer than the medium and maximum clearance cost designs. The maximum clearance cost design achieves the shortest tracking time but the longest path. The medium clearance cost design strikes a balance between them: the path is significantly shorter compared to the maximum clearance cost design while the tracking time is similar. These results suggest that a medium clearance cost design is preferable.

\begin{table}[t]
    \centering
    \caption{Obstacle clearance cost parameters.}
    \label{tab:costmap_design}
    \setlength{\tabcolsep}{10pt}
    \begin{tabular}{|l|c|c|c|} 
        \hline 
        Design &$\kappa$  	&$c_u$   	&$c_f$ \\ 
        \hline
        minimum clearance 	& 15.0 	&3.2  &1 \\ 
        medium  clearance 	& 7.0 	&8.3  &5 \\ 
        maximum clearance 	& 1.0 	&16.9 &15 \\
        \hline
    \end{tabular}
\end{table}

\begin{table*}[t]
    \centering
    \caption{Simulation results for three cost map designs.}
    \label{tab:jackal_race_result}
    \resizebox{\linewidth}{!}{
	\begin{tabular}{|l|c|c|c|c|c|} 
		\hline
		Costmap Design	& Plan Path Length & Robot Traj. Length & Finish Time & Avg. Clearance & Min. Clearance \\ 
		\hline
		minimum clearance   		&15.10 m  &15.66 m 	&32.80 sec	&0.45 m  	&0.13 m \\ 
		medium  \;\,\,clearance  	&16.19 m  &15.63 m  &20.62 sec	&0.63 m		&0.34 m\\ 
		maximum clearance  			&22.28 m  &20.32 m  &18.96 sec  &1.31 m  	&0.38 m\\ 
		\hline
	\end{tabular} 
    }
\end{table*}

\subsection{Adaptive Control Gain Evaluation}
\label{sec:controlgain}

\begin{figure}[t]
    \centering
    \includegraphics[width=0.48\linewidth,valign=t]{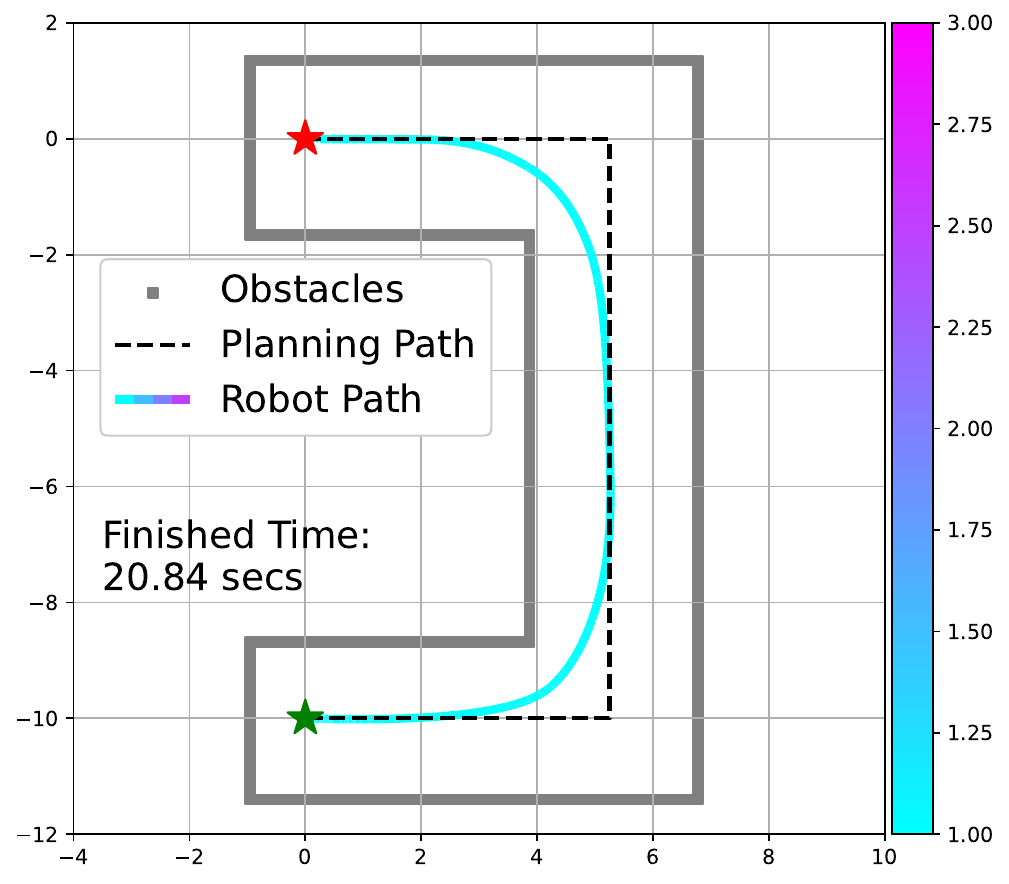}
    \includegraphics[width=0.48\linewidth,valign=t]{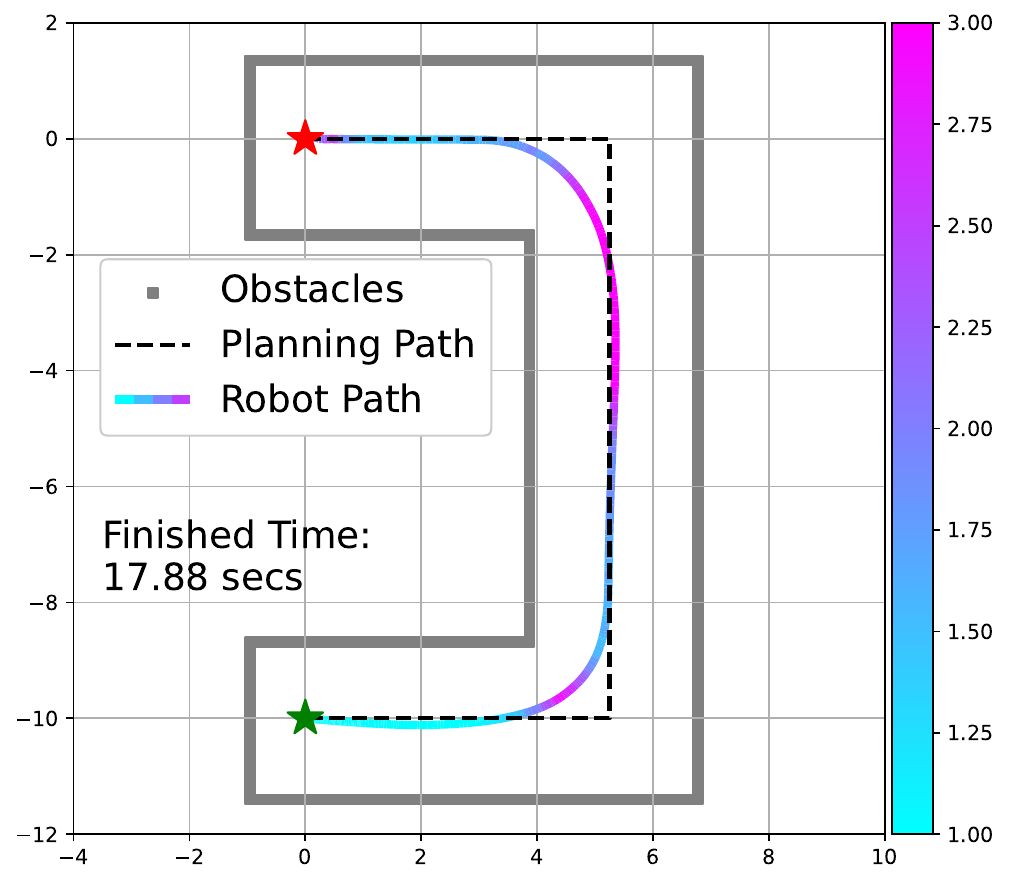}
    \caption{Simulation of our safe tracking controller in a C-shape environment with reference path shown as a black dashed line and start and goal denoted by a red and green star, respectively. The colored curve represents the actual robot trajectory with control gain shown by gradient color corresponding to the side color bar.}
    \label{fig:sddm_boost_comp}
\end{figure}

To demonstrate the effectiveness of the adaptive control gain \eqref{eq:sddm_boost_gain}, we created a C-shaped simulated environment using Gazebo with a pre-specified reference path, shown in Fig.~\ref{fig:sddm_boost_comp}. We compared the performance of the safe tracking controller with fixed control gain $k_v = 1$ versus with the adaptive control gain in \eqref{eq:sddm_boost_gain}. With the adaptive control gain, the robot speeds up faster in straight lines and keeps a low speed in turns, finishing the task in less time compared to fixed control gain design.

\subsection{Baseline Comparison}
\label{sec:eva_planner}

In this section, we compare EAST with the EVA-Planner \citep{eva_planner2021}, which solves the safe tracking problem via two-layer hierarchical MPC. In the first layer, the EVA-Planner refines a planned path utilizing a low-order approximation of the robot's motion model. This refined path is then used to define the cost function in the second layer, which considers tracking error, safety constraints, and dynamics feasibility. All constraints are encoded in the cost function, leading to an unconstrained, nonlinear, non-convex optimization problem, which is solved by a gradient-based numerical solver. We carried out two experiments to compare the methods with emphasis on safety and stability, respectively.

\begin{figure*}[t]
\begin{minipage}[t]{0.32\linewidth}
    \centering
    \includegraphics[width=0.94\linewidth]{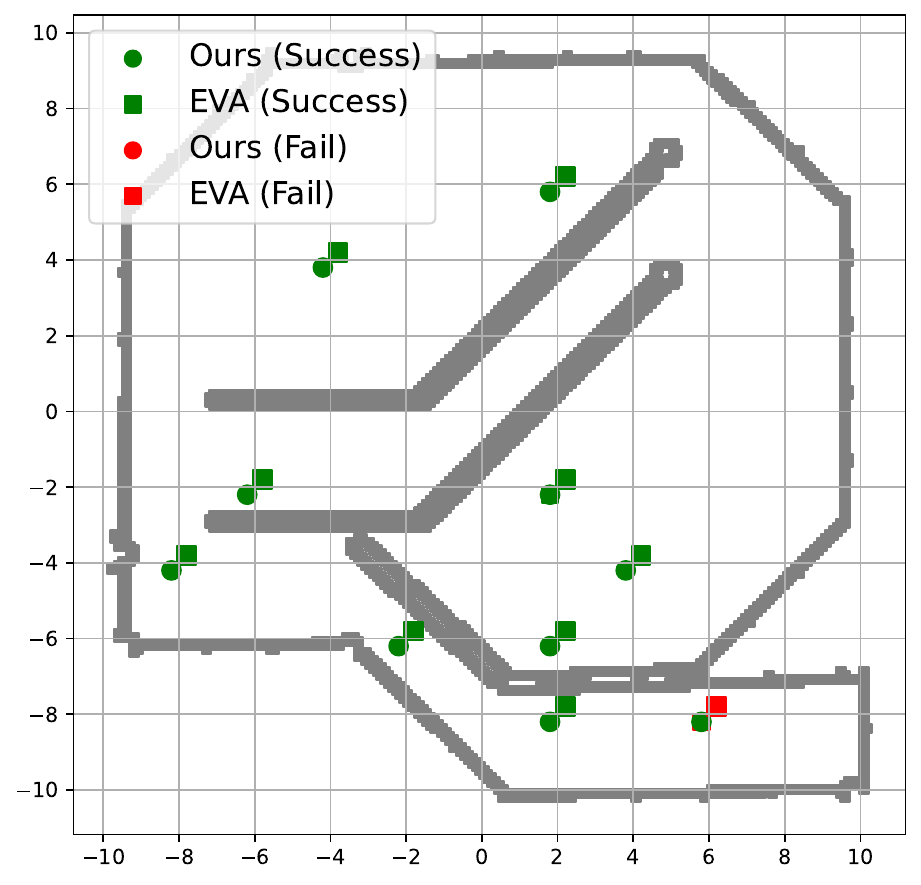}	
    \caption{EAST and EVA-Planner \citep{eva_planner2021} navigating to randomly selected goals in simulation. The successful and unsuccessful goal executions are shown in green and red, respectively.}
    \label{fig:random_ten_safety_test}
\end{minipage}%
\hfill%
\begin{minipage}[t]{0.65\linewidth}
    \includegraphics[width=0.5\linewidth]{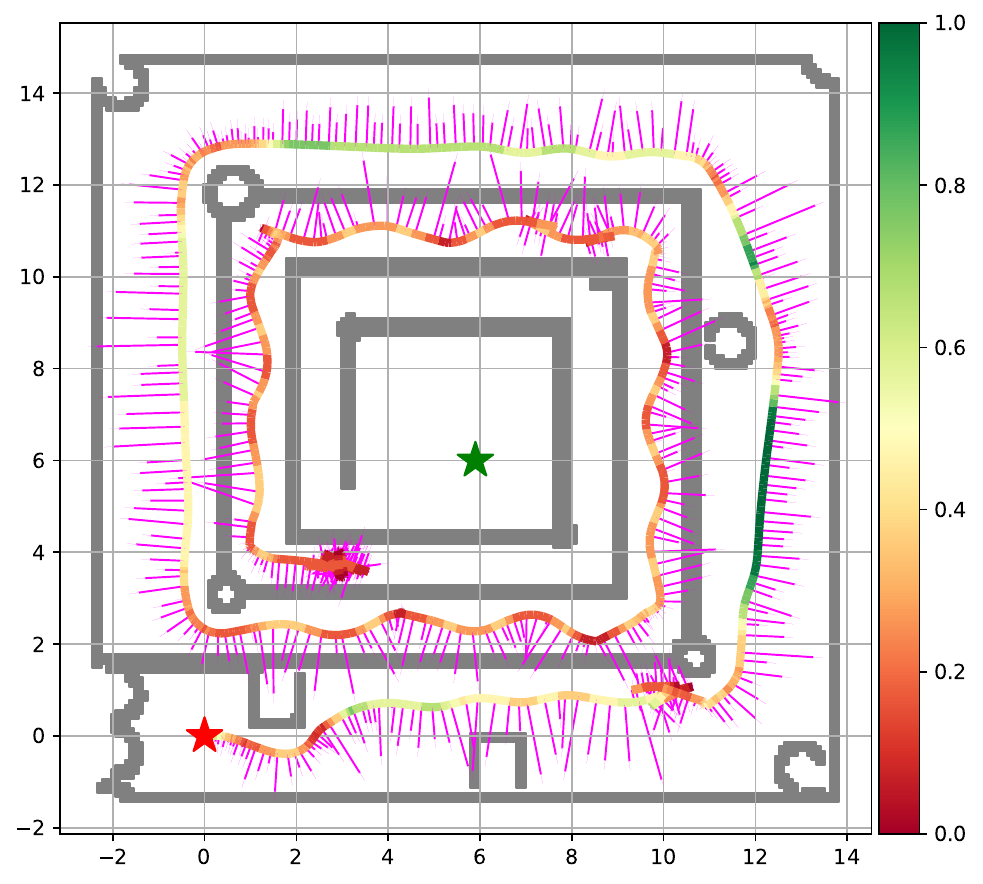}%
    \includegraphics[width=0.5\linewidth]{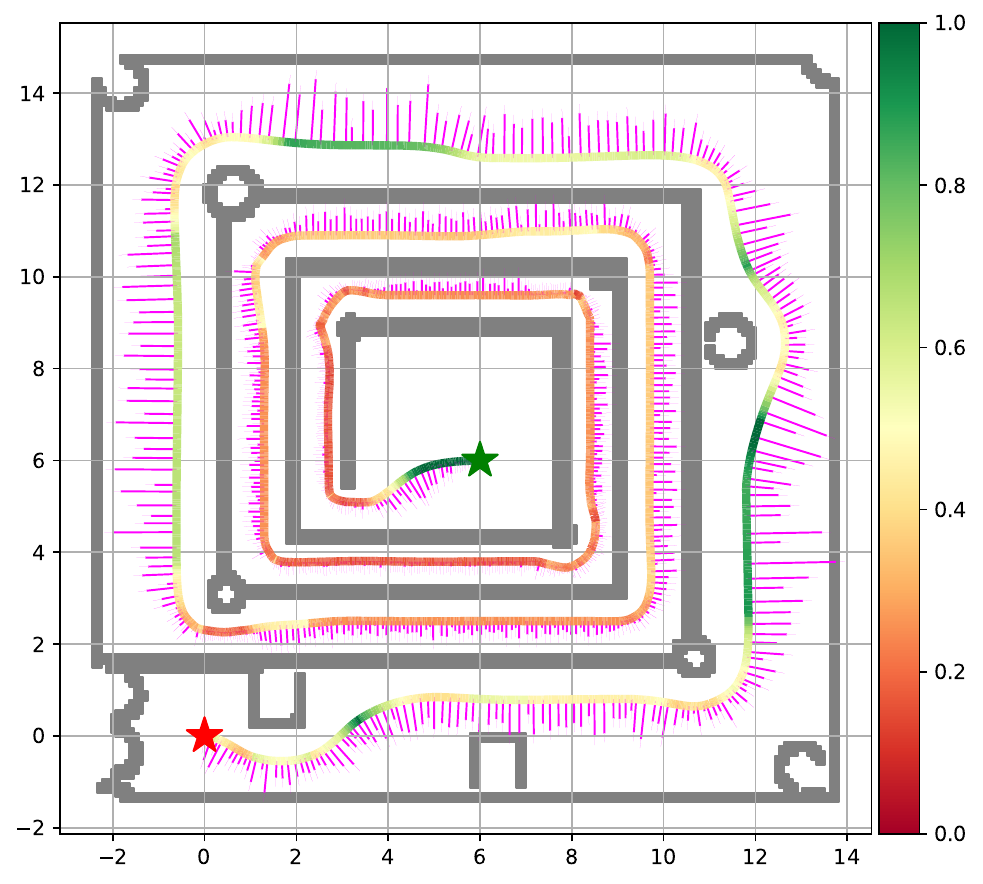}
    \caption{Adaptivity test in the maze environment. The start and the goal are denoted by a red and green star, respectively. The left figure presents the results produced by the EVA-Planner method, while the right figure displays the results obtained using our EAST method. The colored curve represents the actual robot trajectory with the robot to the inflated obstacle space $d(\bfp, \Omega^+)$ shown by gradient color corresponding to the side color bar. The tracking velocity profiles are shown as magenta arrows perpendicular to the robot trajectories.}
    \label{fig:maze_test}
\end{minipage}
\end{figure*}

\textbf{Ten Point Test.} 
This test evaluates the safety of EAST and the EVA-Planner in a static environment. Ten feasible goal points were selected in a simulated Gazebo environment, as shown in Fig.~\ref{fig:random_ten_safety_test}. In each trial, the robot was tasked to reach one of these predetermined goals without any prior knowledge of the environment. The experiment was repeated ten times, so all the goal points were tested.

The results in Fig.~\ref{fig:random_ten_safety_test} show that EAST can reach all ten goals safely, while the EVA-Planner fails to reach one. Since the EVA-Planner encodes the safety constraint in the cost function, solutions to its optimization problem do not guarantee safety, and the output trajectory goes through obstacles during our test. In contrast, EAST guarantees safety via Prop.~\ref{thm:main_result} as long as the planned path $\rho$ is in the interior of the free space and the robot is sufficiently close to the start of the path initially.

\textbf{Maze Test.} 
This test evaluates the stability of EAST and the EVA-Planner in a static environment. A maze-like simulated environment was created, shown in Fig.~\ref{fig:maze_test}, with the objective of going from a start position at the origin to a goal position at $(6, 6)$ in the maze center. To accomplish this task, the robot must go through corridors that become narrower from the outer (about $3$ m wide) to the inner (less than $1$ m wide) part of the environment. 

In Fig.~\ref{fig:maze_test}, the left plot shows the trajectory generated by the EVA-Planner and the right plot shows the trajectory generated by our EAST method. Our method accomplishes this task successfully while adapting its speed according to the local environment geometry, i.e., moving faster when the corridor is wide and slower when it becomes narrower. In contrast, the EVA-Planner cannot handle this task. The robot fails to adjust its speed as the corridor becomes narrower, which leads to oscillatory motion and eventually a safety violation. This oscillatory behavior arises from the gradient-based safety metric employed by the EVA-Planner, which continuously attempts to push the robot away from nearby obstacles. Consequently, within a narrow corridor, when the robot is repelled from one side, it rapidly approaches the opposite side, causing a repeated cycle of oscillations. This back-and-forth behavior persists until a safety violation occurs.

\subsection{Hardware Evaluation in Cluttered Static Environment}
\label{sec:static}
\begin{figure*}[t]
\centering
\begin{subfigure}{0.32\linewidth}
    \centering
    \includegraphics[width=0.95\linewidth]{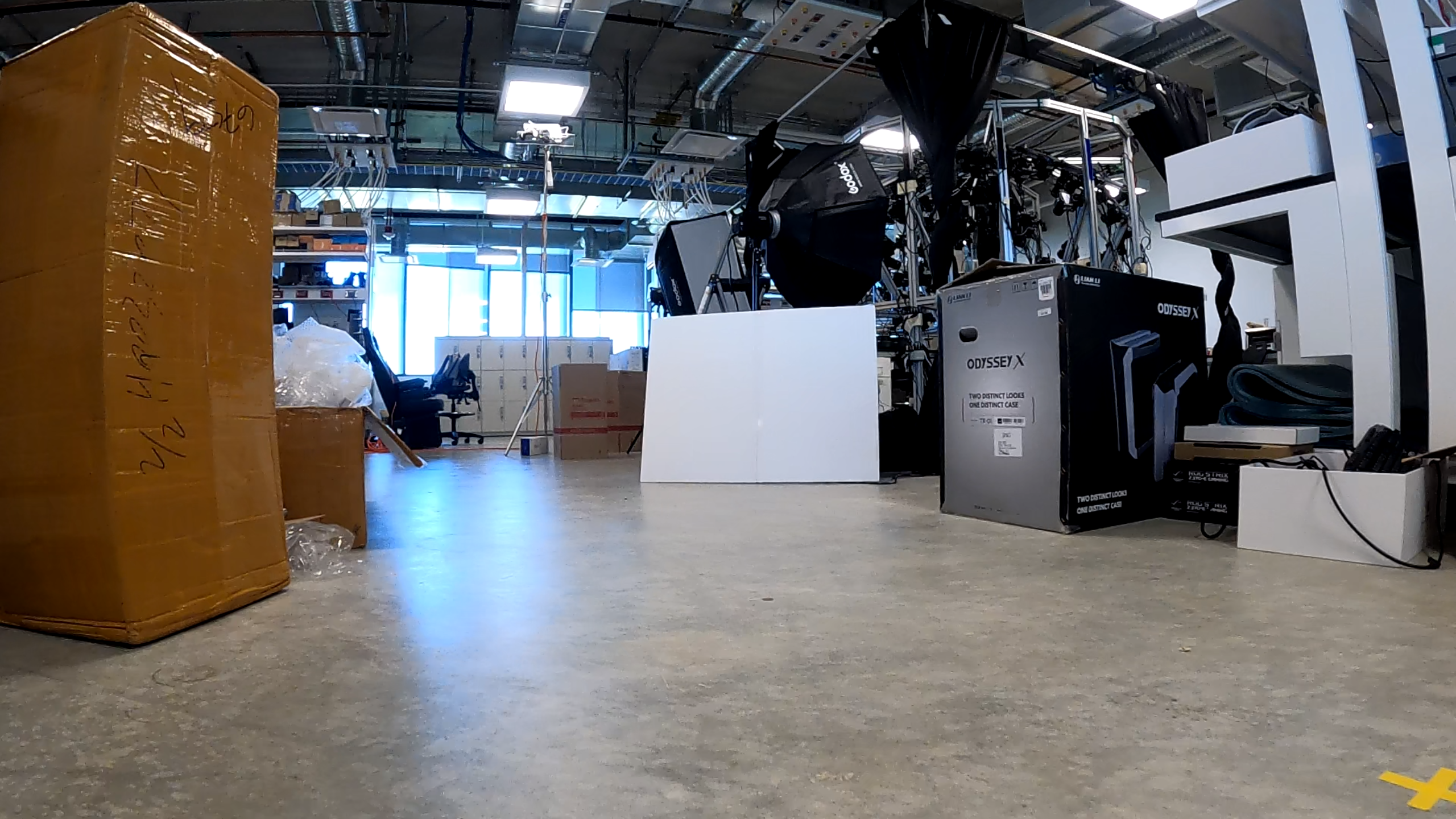}
    \vfill
    \hspace{-0.1in}		
    \includegraphics[width=1\linewidth]{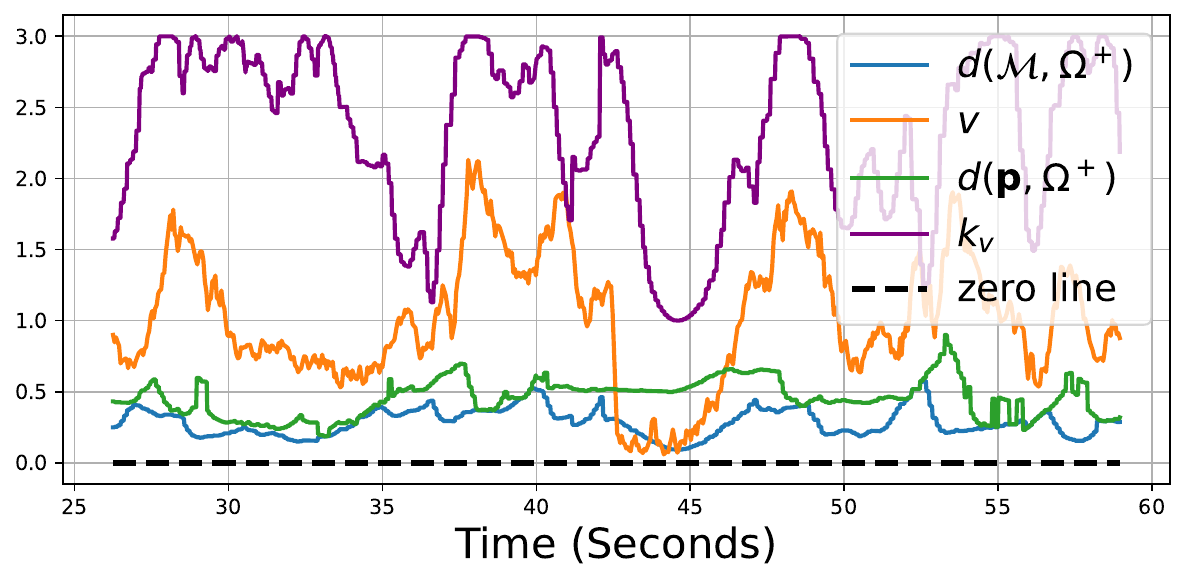}
    \vspace{-0.05in}	
    \caption{Scene 1: Going from wide-open area to cluttered region.}
    \label{fig:fah_scene1}		
\end{subfigure}%
\hfill%
\begin{subfigure}{0.32\linewidth}
    \centering
    \includegraphics[width=0.95\linewidth]{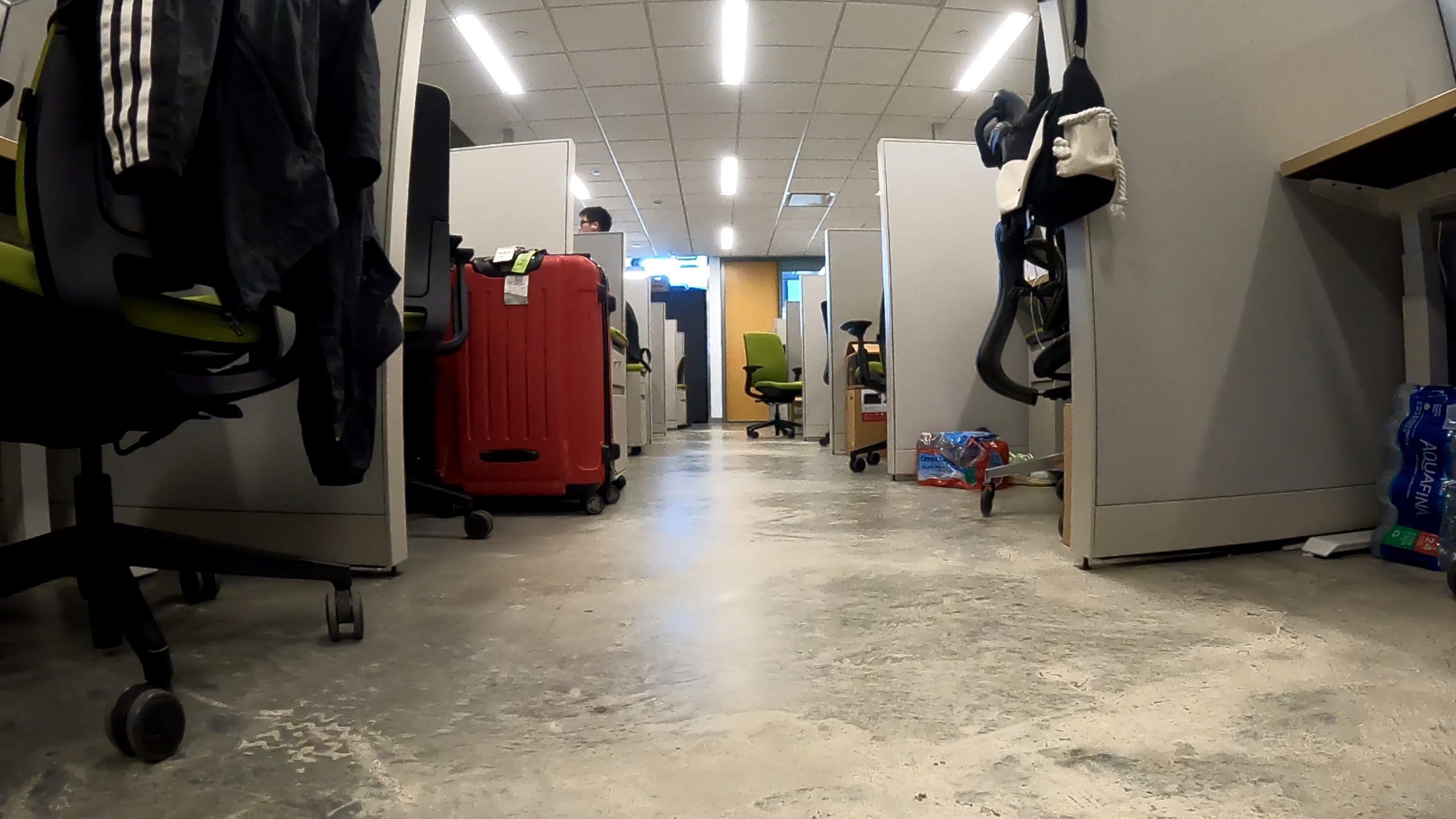}
    \vfill
    \hspace{-0.1in}	
    \includegraphics[width=1\linewidth]{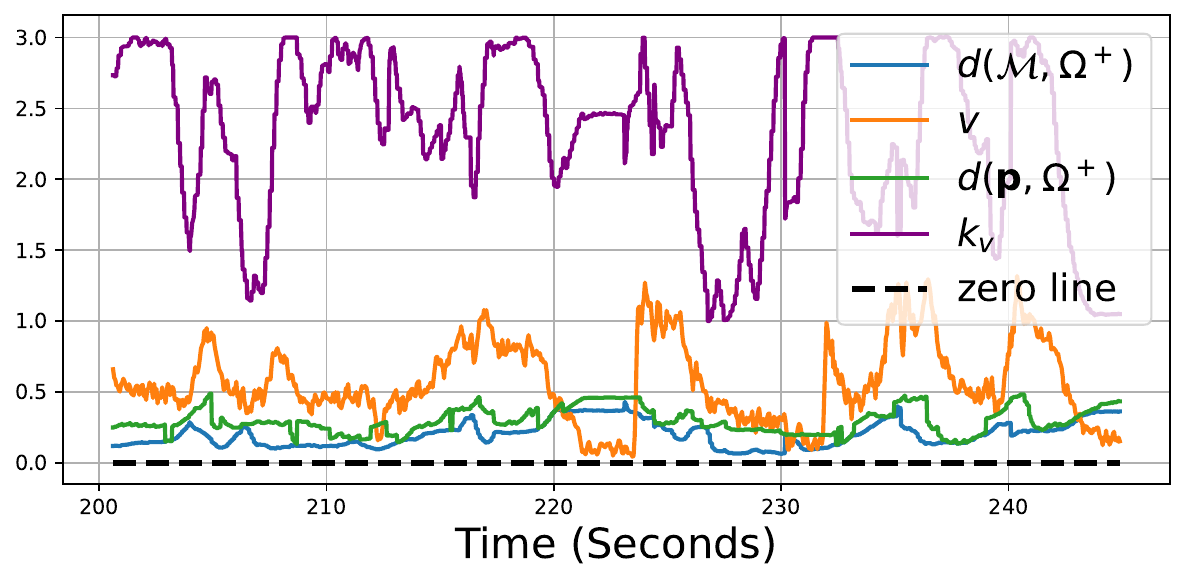}
    \vspace{-0.05in}		
    \caption{Scene 2: Going through narrow aisle with obstacles.}
    \label{fig:fah_scene2}		
\end{subfigure}%
\hfill%
\begin{subfigure}{0.32\linewidth}
    \centering
    \includegraphics[width=0.95\linewidth]{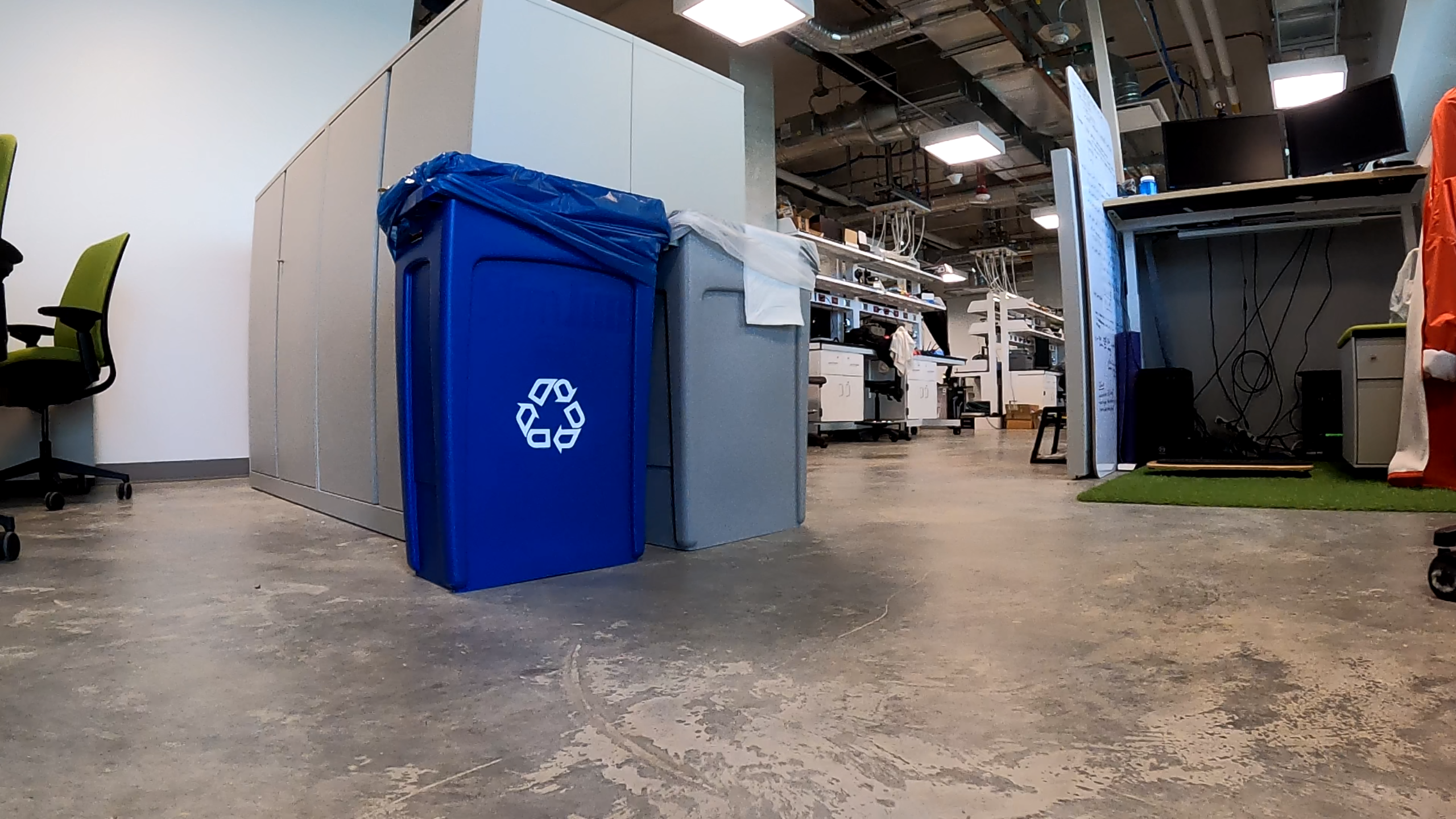}
    \vfill
    \hspace{-0.1in}			
    \includegraphics[width=1\linewidth]{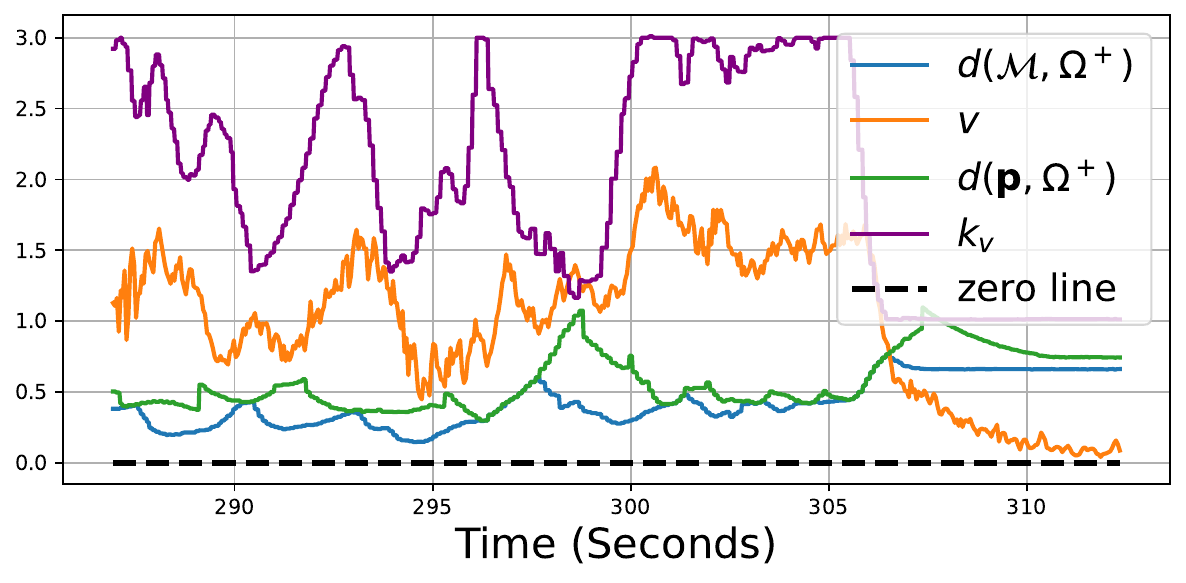}
    \vspace{-0.05in}		
    \caption{Scene 3: Passing from one region to another through narrow gap.}
    \label{fig:fah_scene3}		
\end{subfigure}
\begin{subfigure}{1\linewidth}
    \centering
    \includegraphics[width=\linewidth]{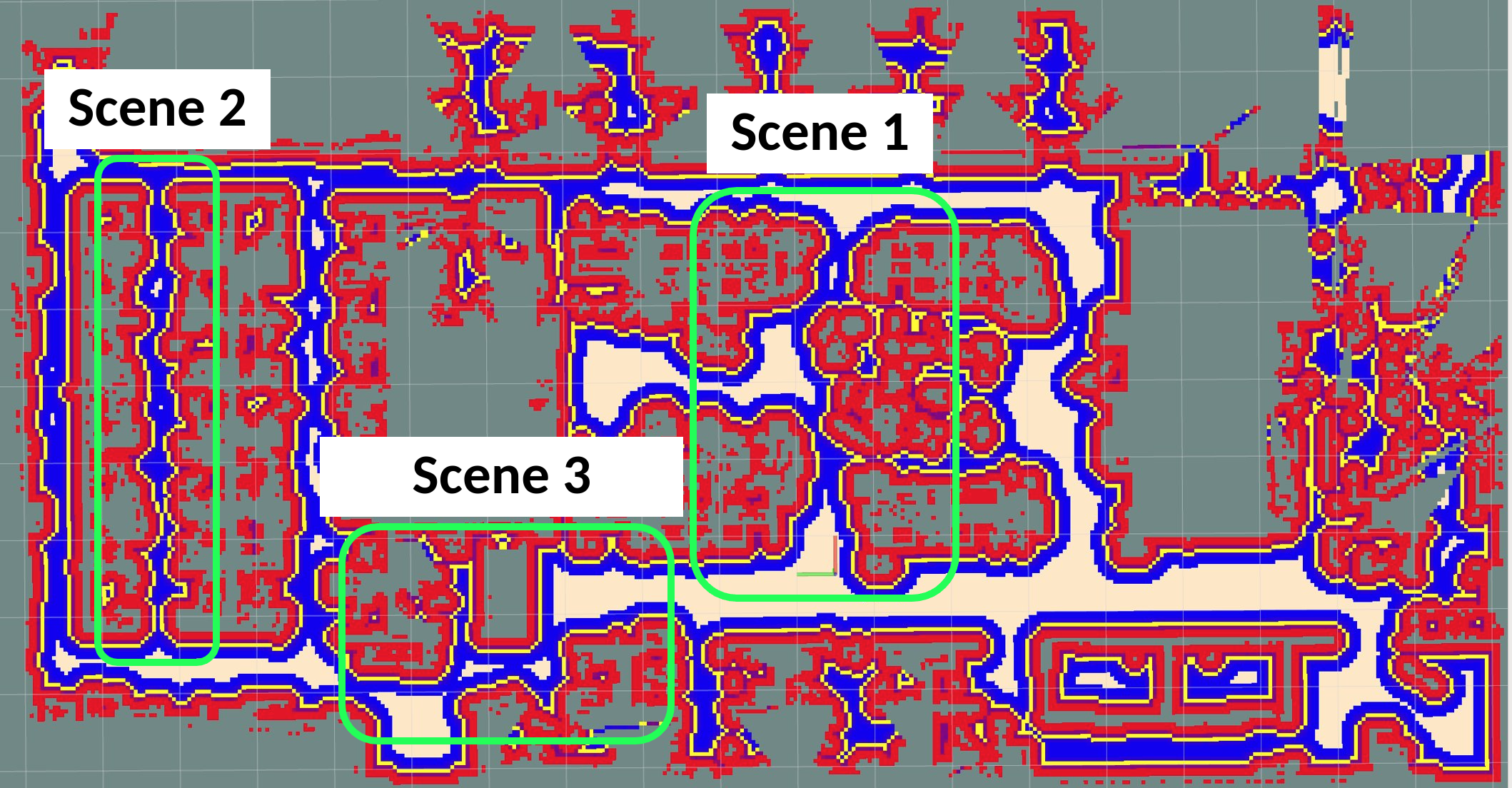}
    \caption{Obstacle clearance cost map with cost values from high to low shown as red, yellow, blue, and white.}
    \label{fig:fah_costmap_with_scene_boxes}
\end{subfigure}
\caption{Hardware experiment in large lab environment (about $40$ m $\times$ $20$ m). Three representative scenes, shown in (a)-(c), are marked on the final obstacle clearance cost map (d) using green regions. The distance from the motion prediction set to the inflated obstacle space $d(\calM,\Omega^+)$, the robot linear velocity $v$, the distance from the robot to the inflated obstacle space $d(\bfp,\Omega^+)$, and the adaptive control gain $k_v$ are plotted beneath each scene.}
\label{fig:fah_exp}
\end{figure*}

This experiment tests the safety and stability of EAST in a real environment of size $40 \times 20$ m, containing static obstacles of different sizes, such as robots, boxes, desks, etc., as shown in Fig.~\ref{fig:fah_scene1}-\ref{fig:fah_scene3}.

During this experiment, a few goals in unknown regions are specified, and the robot is required to navigate autonomously. The robot has no prior knowledge of the environment and is expected to reach each goal without collisions. A first-person video captured by the onboard camera of the robot can be found in the supplementary material. 

The experiment is visualized in Fig.~\ref{fig:fah_exp}, where the final obstacle clearance cost map is depicted in Fig.~\ref{fig:fah_costmap_with_scene_boxes}. Three local scenes (marked in green boxes) are shown in the top row, with associated quantitative results below each of them. The robot navigates adaptively, slowing down when entering obstacle-dense areas and speeding up when in wide open spaces, as shown in Fig.~\ref{fig:fah_scene1}-\ref{fig:fah_scene3}. From Fig.~\ref{fig:fah_scene2}, we can see that the speed of the robot (orange curve) is higher when the distance to the inflated obstacle space (green line) is larger, and the adaptive control gain (purple curve) increases when the robot heading is aligned with the local environment. 

\subsection{Evaluation in Dynamic Environments}
\label{sec:dynamic}

This section evaluates the safety and stability of EAST in dynamically changing environments. The robot is assumed to know the position and velocity of the moving obstacles. In the simulated experiments, the trajectories of the moving obstacles were pre-specified with position and velocity known to the robot. In the hardware experiments, a Vicon motion capture system was used to track the position and velocity of human actors and transmit the motion information to the robot using a wireless network. In practice, this information can be obtained from object detection and tracking using visual or LiDAR sensors. The control parameters used in this section are summarized in Tab.~\ref{tab:ctrl_params_mov_sim}.

\begin{table}[t]
    \centering
    \caption{Control gain parameters used in the dynamic environment simulations.} \label{tab:ctrl_params_mov_sim}
    \begin{tabular}{ |l|c|c|c|c|c|c| } 
        \hline 
        Parameter 	&$k_g$ & $k_v$ & $k_\omega$ & $\gamma_i$  &	$q_1$	& $q_2$ \\ 
        \hline
        Value 		&$2.0$ & $2.0$ 	&$5.0$	&$0.15$	 &$1$		& $1$  \\ 
        \hline
    \end{tabular}
\end{table}

\begin{figure*}[t]
	\centering
	\begin{subfigure}{0.32\textwidth}
		\centering
		\includegraphics[width=1\textwidth]{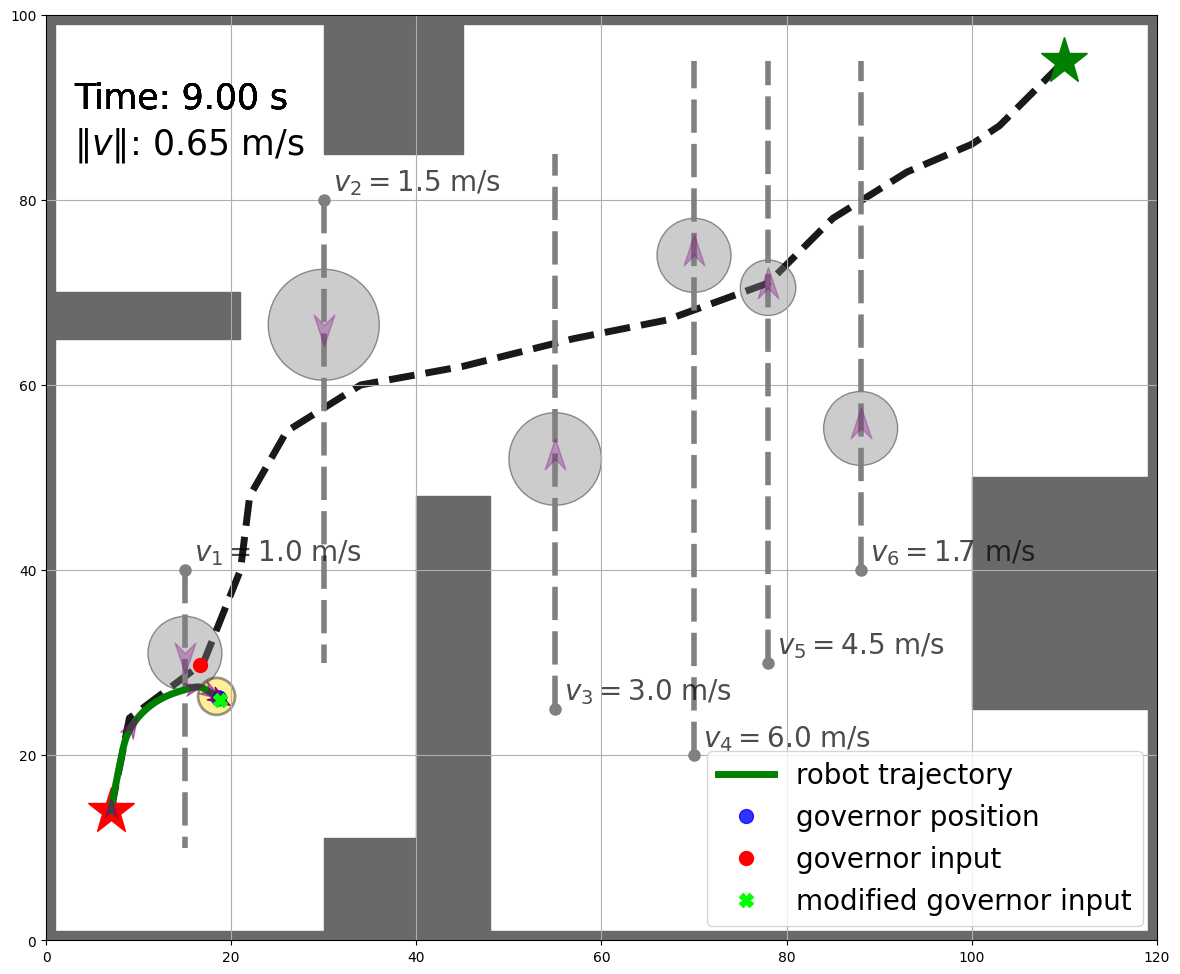}
		\caption{snapshot 1}
		\label{fig:mov_exp_sim:snapshot1}	
	\end{subfigure}%
	\hfill%
	\begin{subfigure}{0.32\textwidth}
		\centering
		\includegraphics[width=1\textwidth]{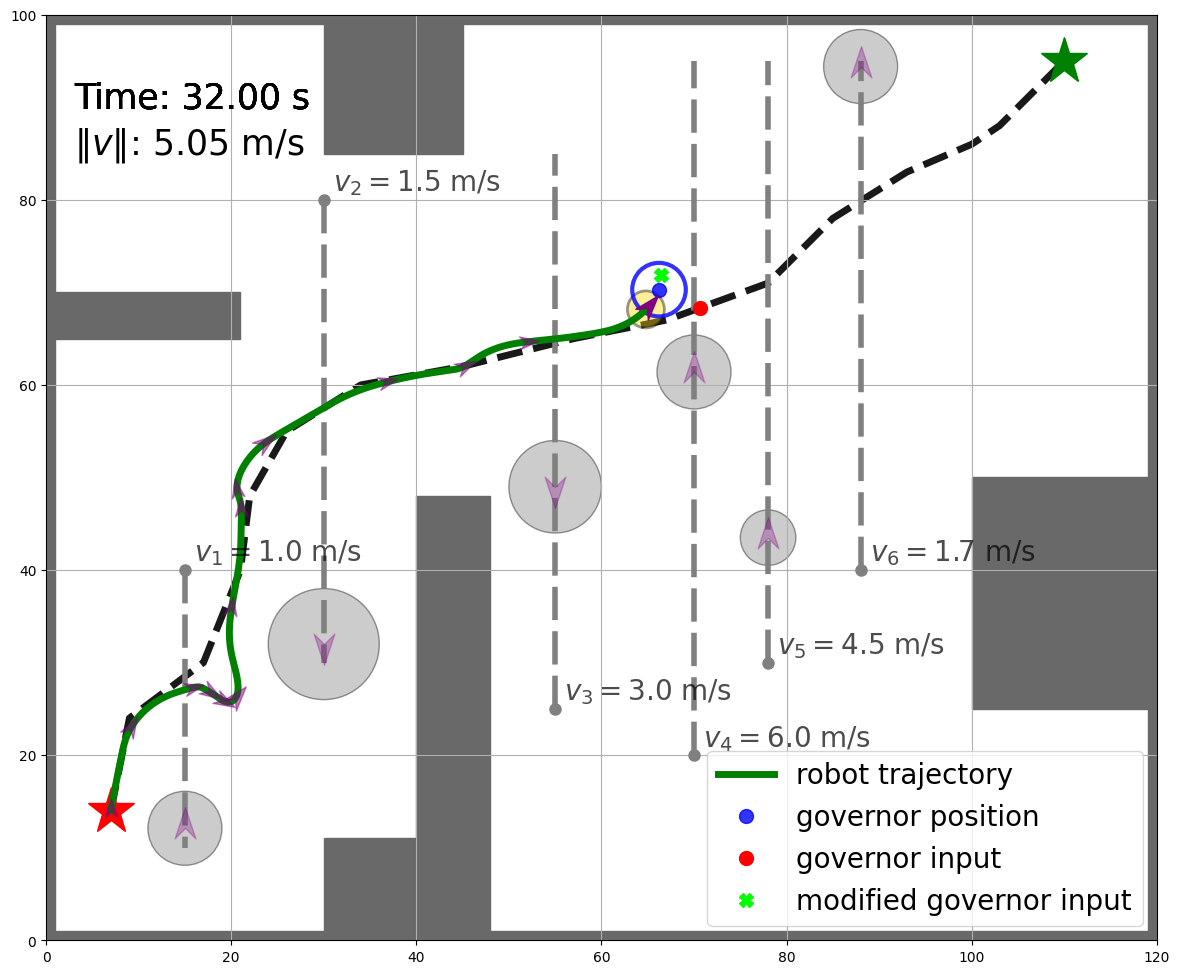}
		\caption{snapshot 2}
		\label{fig:mov_exp_sim:snapshot2}		
	\end{subfigure}%
	\hfill%
	\begin{subfigure}{0.32\textwidth}
		\centering
		\includegraphics[width=1\textwidth]{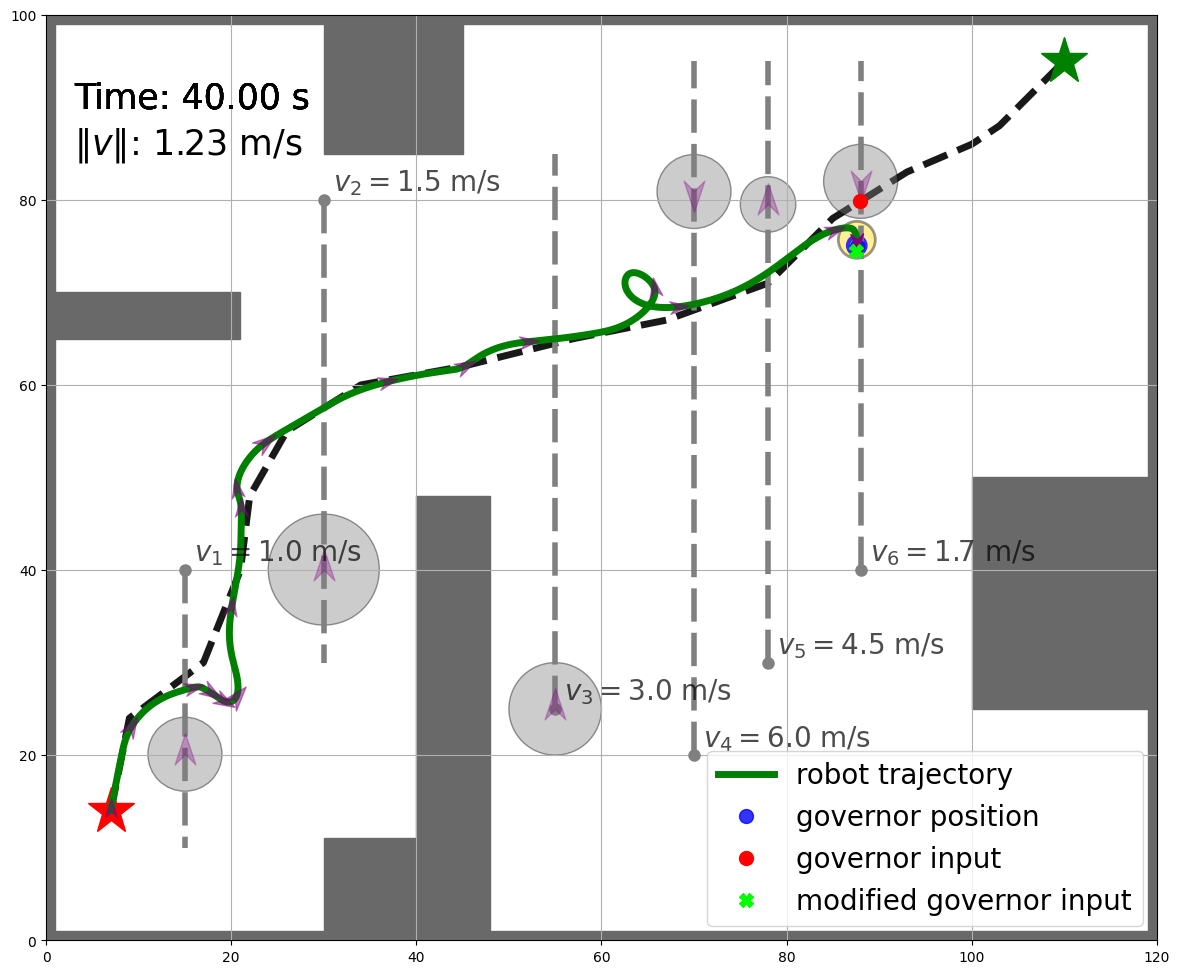}
		\caption{snapshot 3}
		\label{fig:mov_exp_sim:snapshot3}		
	\end{subfigure}
	\begin{subfigure}{1\textwidth}
		\centering
		\includegraphics[width=\textwidth]{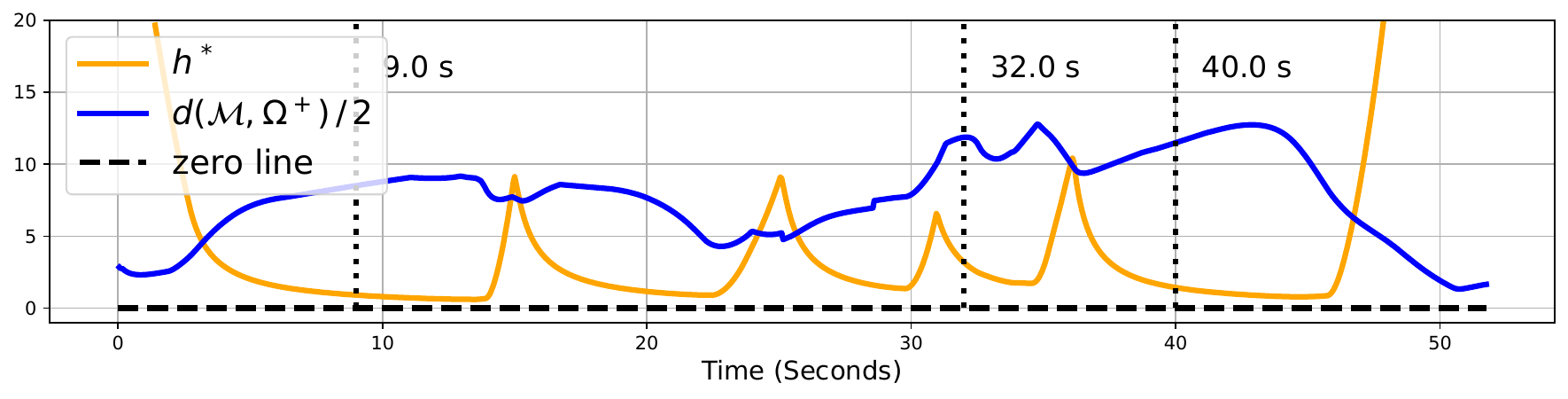}
		\caption{Distance from motion prediction set to inflated obstacles $d(\calM,\Omega^+)$ and minimum CBF value $h^* = \min_{i} h_i$.}
		\label{fig:mov_exp_sim:safety_metric}
	\end{subfigure}
	\caption{Snapshots (a)-(c) from simulation with six dynamic obstacles with different sizes and speeds, shown as gray circles. The pre-specified paths for the moving obstacles are depicted using gray dashed lines. The start and goal locations for the robot are shown by a red and green star, respectively. The robot is represented as a yellow circle. The robot's reference path is shown as a black dashed line. The actual robot trajectory is shown as a green solid line with purple arrows indicating its heading. The governor position $\bfg$, governor input $\bfu_g$, and modified governor input $\bar{\bfu}_g$ are shown by blue, red, and green dots, respectively.}
    \label{fig:mov_exp_sim}
\end{figure*}

\textbf{Simulation experiment.} 
We evaluated EAST in a simulated dynamically changing environment, containing four static obstacles and six moving obstacles (light gray circles) of different sizes and velocities, shown in Fig.~\ref{fig:mov_exp_sim:snapshot1}-\ref{fig:mov_exp_sim:snapshot3}. Each moving obstacle follows a pre-specified path (gray dashed lines). The robot is required to reach a goal location without collisions with either static or moving obstacles.

The results of this experiment can be found in Fig.~\ref{fig:mov_exp_sim}. The robot is able to reach the goal successfully without collisions. The three snapshots in Fig.~\ref{fig:mov_exp_sim} show interactions between the robot and the moving obstacles. The robot deviates from its planned trajectory to avoid collisions with the moving obstacles effectively and subsequently returns to its original path once the obstructions are no longer present.

\begin{figure*}[t]
    \centering
    \begin{subfigure}{0.69\linewidth}
        \centering
        \includegraphics[width=1\linewidth]{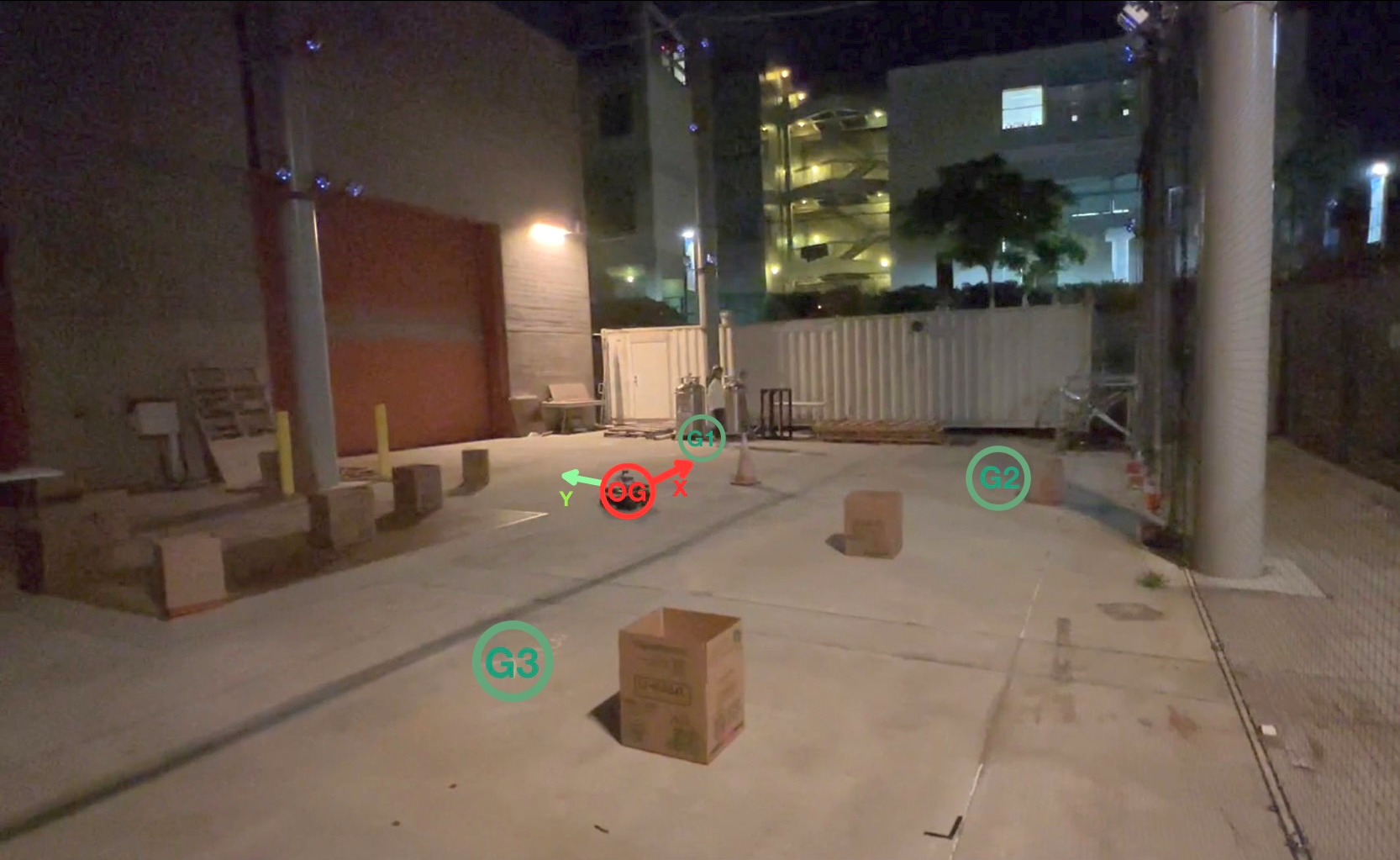}
        \caption{Experiment site with Vicon motion capture system}
        \label{fig:mov_exp_hw:aerodrome}		
    \end{subfigure}%
    \hfill%
    \begin{subfigure}{0.31\linewidth}
        \centering
        \includegraphics[width=0.93\linewidth]{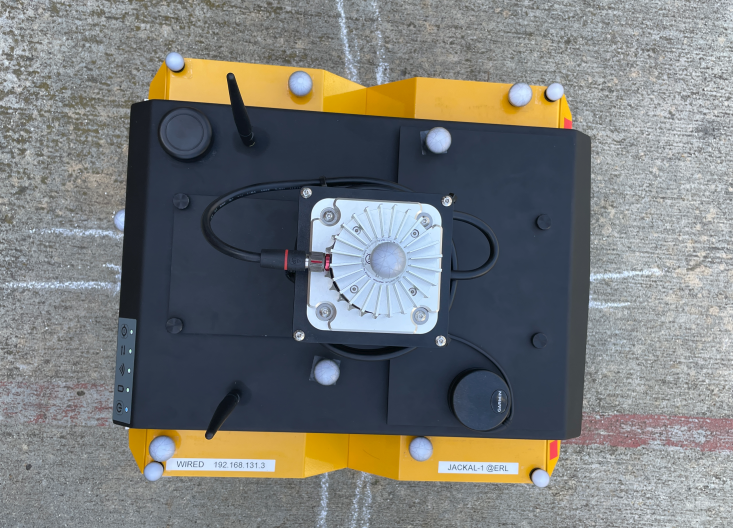}\\
        \vfill
        \vspace*{0.05cm}
        \includegraphics[width=0.93\linewidth]{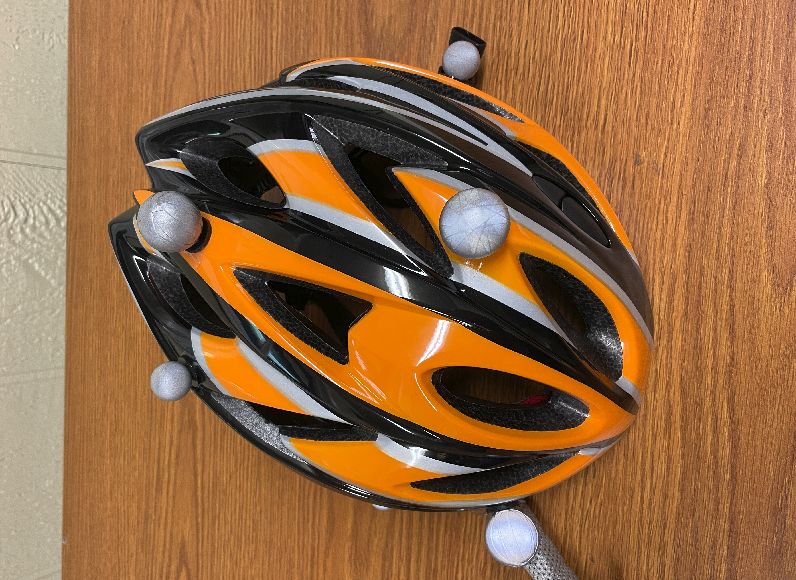}
        \caption{Robot and helmet with markers}
        \label{fig:mov_exp_hw:jackal}		
    \end{subfigure}
    \begin{subfigure}{1\linewidth}
        \centering	
        \includegraphics[width=\linewidth,trim={0 10pt 0 0},clip]{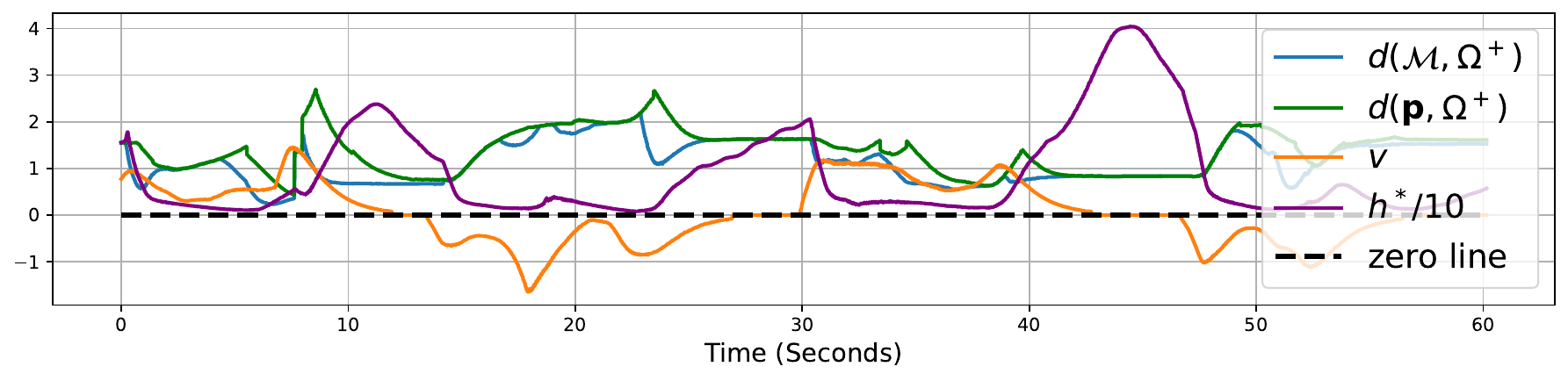}
        \caption{Distance from motion prediction set to inflated obstacles $d(\calM,\Omega^+)$, distance from robot to inflated obstacles $d(\bfp,\Omega^+)$, robot linear velocity $v$, and minimum CBF value $h^* = \min_{i} h_i$.}
        \label{fig:mov_exp_hw:safety_metric}
    \end{subfigure}
    \caption{Hardware experiment with Jackal robot and two pedestrians, serving as moving obstacles. The actors' positions are captured by a Vicon motion capture system and passed to the robot over a WiFi network. The robot re-plans its path at 10 Hz online to reach different goals while avoiding the pedestrians crossing its path. The three goals are marked as green circles, and the origin is marked as a red circle in (a). The robot and helmets worn by the pedestrians are shown in (b). Quantitative results from the experiment are shown in (c). Please refer to the supplementary material for a video from the experiment.
    }
    \label{fig:mov_exp_hw}
\end{figure*}

\textbf{Hardware experiment.} 
Finally, we evaluated the safety and stability of EAST in a real, dynamically changing environment. The experiment setup is shown in Fig.~\ref{fig:mov_exp_hw} with static obstacles (boxes, walls, pillars, etc.) and moving obstacles (two human pedestrians wearing helmets). The robot does not have prior information about this environment, except for receiving moving obstacle positions and velocities from the Vicon system. The robot is required to visit three pre-specified goal states and return to the start state without any collision.

The quantitative results from this experiment are shown in Fig.~\ref{fig:mov_exp_hw:safety_metric}. We observe that our method successfully executes the experiment. During the test, the planner failed to adapt to the pedestrians due to its low re-planning frequency, and the robot made multiple deviations from the planned path to avoid moving obstacles. This stems from the planner's dependence on an occupancy map, which is not updated sufficiently fast to capture changes in a timely manner. Consequently, safe navigation in dynamic settings cannot rely solely on the planned path. The control algorithm plays a critical role in departing from the nominal path whenever strict adherence would place the robot in unsafe states.

Both the simulation and hardware experiments demonstrate that our EAST method is able to successfully handle static and dynamic obstacles by optimizing the reference governor behavior in real time. Our QCQP optimization of the reference governor input with CBF constraints capturing the dynamic obstacles demonstrates several advantages. First, imposing the CBF constraint on the simpler reference‑governor dynamics substantially lowers the computational load of the online optimization, enabling real‑time adaptation to moving obstacles. Second, the reference signal is altered only when its direct execution would compromise safety, thereby preserving the nominal stability of the original reference governor formulation with minimal degradation. Moreover, by decoupling path‑tracking control from safety enforcement and confining the complex safety constraints to the reference‑governor dynamics, our formulation is readily transferable to robot platforms with different system dynamics.

\section{Conclusion}
\label{sec:conclusion}
This paper presented an environment-aware safe tracking controller using a reference governor design to enforce safety in dynamic environments. Our control design separates path tracking from safety constraint enforcement using a reduced-order reference governor system. Using prior results, we defined a local safe zone for the robot-governor system as the set of governor inputs where the reachable set of the robot remains outside the obstacle set. While choosing governor inputs in the local safe zone guarantees static obstacle safety, our key contribution is to optimize the governor input to also satisfy CBF constraints capturing dynamic obstacles. The resulting convex QCQP optimization can be solved in real-time, enabling fast adaptation in dynamic environments. The method guarantees safety in static environments, while in dynamic environments safety is maintained only when the QCQP is feasible. We analyzed the feasibility of the optimization problem in Sec.~\ref{sec:ref_gvn_extension} and discussed how our method can be applied to a broader class of systems in Sec.~\ref{sec:generalization}. Our controller outperformed a baseline method in simulation and demonstrated safety, efficiency, and fast adaptation to moving obstacles in several real environments. Future work will focus on extending our adaptive reference governor design to robots with more complex dynamics, such as mobile manipulators.

\section*{Funding}
We gratefully acknowledge support from ONR N00014-23-1-2353 and NSF RI IIS-2007141.

\section*{Statements and Declarations}
\textbf{Conflict of interest} \ The authors have no competing interests to declare that are relevant to the content of this article.

\bigskip

\noindent \textbf{Author contribution} \ Conceptualization: ZL, NA; Methodology: ZL, YY, ZN, NA; Theoretical analysis: ZL, YY, ZN, NA; Algorithm development: ZL, YY, ZN, NA; Simulation validation and hardware evaluation: ZL, YY, ZN; Writing - initial draft: ZL, YY; Writing - review and editing: YY, NA; Supervision: NA.

\bibliography{bib/ref.bib}

\end{document}